\documentclass{article}

\usepackage{microtype}
\usepackage{subfigure}
\usepackage{url}
\usepackage{booktabs} 
\usepackage{cancel}

\newcommand{\vlat}{\vz}
\newcommand{\lat}{z}

\newcommand{\varpar}{\lambda}
\newcommand{\vvarpar}{\vlambda}

\newcommand{\mix}{w}

\newcommand\cut[1]{}

\newcommand{\squishlist}{
   \begin{list}{$\bullet$}
    { \setlength{\itemsep}{0pt}      \setlength{\parsep}{3pt}
      \setlength{\topsep}{3pt}       \setlength{\partopsep}{0pt}
      \setlength{\leftmargin}{1.5em} \setlength{\labelwidth}{1em}
      \setlength{\labelsep}{0.5em} } }

\newcommand{\squishlisttwo}{
   \begin{list}{$\bullet$}
    { \setlength{\itemsep}{0pt}    \setlength{\parsep}{0pt}
      \setlength{\topsep}{0pt}     \setlength{\partopsep}{0pt}
      \setlength{\leftmargin}{2em} \setlength{\labelwidth}{1.5em}
      \setlength{\labelsep}{0.5em} } }

\newcommand{\squishend}{
    \end{list}  }

{}
{}
{}
{}
\newtheorem{defn}{Definition}{}

\newcommand{\half}{\mbox{$\frac{1}{2}$}}

\newcommand{\sqr}[1]{\left[#1\right]}
\newcommand{\crl}[1]{\left\{#1\right\}}
\newcommand{\myang}[1]{\langle#1\rangle}
\newcommand{\myexpect}{\mathbb{E}}
\newcommand{\Unmyexpect}[1]{\mathbb{E}_{\scaleto{#1\mathstrut}{6pt}}}

\newcommand{\expdist}{\mbox{Exp}}

\newcommand{\gauss}{\mbox{${\cal N}$}}

\newcommand{\IG}{\mbox{IG}}
\newcommand{\IGauss}{\mbox{InvGauss}}

\newcommand{\myvec}[1]{\mbox{$\mathbf{#1}$}}
\newcommand{\myvecsym}[1]{\mbox{$\boldsymbol{#1}$}}

\newcommand{\valpha}{\mbox{$\myvecsym{\alpha}$}}

\newcommand{\vmu}{\mbox{$\myvecsym{\mu}$}}
\newcommand{\vlambda}{\mbox{$\myvecsym{\lambda}$}}

\newcommand{\vphi}{\mbox{$\myvecsym{\phi}$}}

\newcommand{\vPsi}{\mbox{$\myvecsym{\Psi}$}}

\newcommand{\vSigma}{\mbox{$\myvecsym{\Sigma}$}}

\newcommand{\ve}{\mbox{$\myvec{e}$}}

\newcommand{\vf}{\mbox{$\myvec{f}$}}

\newcommand{\vh}{\mbox{$\myvec{h}$}}

\newcommand{\vt}{\mbox{$\myvec{t}$}}
\newcommand{\vu}{\mbox{$\myvec{u}$}}

\newcommand{\vz}{\mbox{$\myvec{z}$}}

\newcommand{\vT}{\mbox{$\myvec{T}$}}

\newcommand{\be}{\begin{equation}}
\newcommand{\ee}{\end{equation}}
\newcommand{\bea}{\begin{eqnarray}}
\newcommand{\eea}{\end{eqnarray}}
\newcommand{\beaa}{\begin{eqnarray*}}
\newcommand{\eeaa}{\end{eqnarray*}}

\usepackage[accepted]{icml2019}
\usepackage{scalerel}
\usepackage{graphicx,color}
\usepackage{amsmath}
\usepackage{verbatim}
\usepackage{amsfonts}

\newtheorem{theorem}{Theorem}
\newtheorem{corollary}{Corollary}[theorem]
\newtheorem{example}{Example}[theorem]

\newtheorem{lemma}{Lemma}

\newenvironment{proof}{\paragraph{Proof:}}{\hfill$\square$}

\newtheorem{property}{Property}{}

\icmltitlerunning{ Stein's Lemma for the Reparameterization Trick}

\begin{document}

\twocolumn[
\icmltitle{
Stein's Lemma for the Reparameterization Trick \\
with Exponential Family Mixtures
}




\begin{icmlauthorlist}
\icmlauthor{Wu Lin}{ubc}
\icmlauthor{Mohammad Emtiyaz Khan}{riken}
\icmlauthor{Mark Schmidt}{ubc}
\end{icmlauthorlist}

\icmlaffiliation{ubc}{University of British Columbia, Vancouver, Canada.}
\icmlaffiliation{riken}{RIKEN Center for Advanced Intelligence Project, Tokyo, Japan}

\icmlcorrespondingauthor{Wu Lin}{yorker.lin@gmail.com}

\icmlkeywords{Gradient Estimation, Stein's Lemma, Gaussian Mixture, Statistics, Machine Learning, ICML}

\vskip 0.3in
]



\printAffiliationsAndNotice{}  

\begin{abstract}
Stein's method \citep{stein1973estimation,stein1981estimation}
 is a powerful tool for statistical applications and has significantly impacted machine learning. Stein's lemma plays an essential role in Stein’s method.
Previous applications of Stein's lemma either required strong technical assumptions or were limited to Gaussian distributions with restricted covariance structures. 
In this work, we extend Stein's lemma to
exponential-family mixture distributions, including Gaussian distributions with full covariance
structures. Our generalization enables us to establish a connection between Stein's lemma and
the reparameterization trick to derive gradients of
expectations of a large class of functions under
weak assumptions. Using this connection, we
can derive many new reparameterizable gradient identities that go beyond the reach of existing
works. For example, we give gradient identities when the expectation is taken with respect to
Student's t-distribution, skew Gaussian, exponentially modified Gaussian, and normal inverse
Gaussian.

\end{abstract}

\section{Introduction}
Stein's lemma \citep{stein1973estimation,stein1981estimation,liu1994siegel} plays an essential role in Stein's method.
The lemma gives a first-order identity to estimate the mean of a multivariate Gaussian distribution.
In machine learning, \citet{fan2015fast,erdogdu2015newton,rezende2014stochastic} use integration by parts to extend the lemma without giving technical conditions.
Another applications of Stein's lemma are De Bruijn's identity \citep{park2012equivalence} and the heat equation identity \citep{brown2006heat}.
These two works give the same second-order identity to estimate the covariance of a multivariate Gaussian distribution.
The second-order identity gives a better unbiased estimation than the one obtained from the first-order identity in terms of variance \citep{khan2017conjugate,khan18a,salimans2013fixed}.
However, \citet{park2012equivalence,brown2006heat} use stronger assumptions to simplify proofs where the authors either assume diagonal covariance structure or twice continuous differentiability.
In gradient estimation, the first-order identity is known as Bonnet's theorem \citep{bonnet1964transformations}.
Bonnet's theorem gives the reparameterization gradient for the mean \citep{kingma2013auto}.
The second-order identity is known as Price's theorem \citep{price1958useful}.
However, \citet{price1958useful,bonnet1964transformations} use characteristic functions as the proof technique. This technique is not easy to extend to a Gaussian mixture and be used to identify weak assumptions.
Beyond Gaussian distribution, Stein's lemma is proposed by \citet{hudson1978natural,brown1986fundamentals}.
In machine learning, \citet{salimans2013fixed,figurnov2018implicit} propose an implicit reparameterization trick under continuous differentiability for a class of distributions.
In fact, the implicit reparameterization gradient can be obtained by Stein's lemma under weaker assumptions.

In this work, we generalize Stein's lemma to Gaussian variance-mean mixture with arbitrary covariance structure and exponential family mixtures while keeping assumptions weak and proofs simple. 
Our theory also directly connects Stein's lemma and the reparameterizable gradient estimation.
Moreover, we present a second-order identity for the covariance estimation of the mixture.
Last but not least, we show that we can obtain the implicit reparameterization trick via Stein's lemma under weaker assumptions than 
\citet{salimans2013fixed,figurnov2018implicit}.
Last but not least, we give examples of gradient identities derived from our theory, such as multivariate Student's t distribution, multivariate skew Gaussian, multivariate exponentially modified Gaussian, and multivariate normal inverse Gaussian.
We find out that these identities are useful in variational inference with Gaussian variance-mean mixture approximations, as shown in \citet{lin2019fast}.

\section{Related Works}
There are many existing works about Stein's lemma. 
\citet{stein1973estimation,stein1981estimation} give a first-order identity for a multivariate Gaussian with diagonal covariance structure.
\citet{liu1994siegel} extends the first-order identity to a multivariate Gaussian with arbitrary covariance structure.
For gradient estimation, Stein’s lemma indeed recovers Bonnet's theorem \citep{bonnet1964transformations}, which is now known as the reparameterization trick \citep{kingma2013auto} with respect to the mean. 
Price's theorem \citep{price1958useful} gives the second-order identity for a multivariate Gaussian with arbitrary covariance structure.
However, \citet{price1958useful,opper2009variational} use the characteristic function of Gaussian to prove Price's theorem, which is not easy to extend to the Gaussian mixture case.
\citet{hudson1978natural,brown1986fundamentals,arnold2001multivariate,landsman2006generalization,landsman2008stein,kattumannil2009stein,kattumannil2016generalized,adcock2007extensions,adcock2012multivariate} further extend Stein's lemma to exponential family and beyond.
Unfortunately, these
works neither show the connection between the gradient
identity and the implicit reparameterization trick nor give
any second-order identity


\section{Smoothness Assumptions}
We first give smoothness conditions. These conditions will be used in the gradient identities.
A key definition is the absolute continuity (AC) of a function: $\vh(\cdot): [a,b] \mapsto \mathcal{R}^m$, where $[a,b]$ is a compact interval in $\mathcal{R}$.
\begin{defn}
A vector function: $\vh(\cdot): [a,b] \mapsto \mathcal{R}^m$ is AC if the following assumptions are satisfied.
\begin{itemize}
 \item Its derivative $\nabla_{\lat} \vh(\lat)$ exists almost everywhere for $\lat \in [a,b]$.
 \item The derivative is Lebesgue integrable. In other words, $\int_{a}^{b} \left|\left| \nabla_{\lat}  \vh(\lat) \right|\right| d\lat <\infty$, where $||\cdot||$ denotes the Euclidean norm.
 \item The fundamental theorem of calculus holds, that is, $\vh(\lat) = \vh(a) + \int_{a}^{\lat} \nabla_{\lat} \vh(t) dt $ for any $\lat \in [a,b]$.
\end{itemize}
\end{defn}
Since we want to deal with a class of functions whose domain is $\mathcal{R}$, we define the locally AC of this class of functions.
\begin{defn}
Let $\vh(\cdot):\mathcal{R} \mapsto \mathcal{R}^m$ be a vector function.
If $\vh(\cdot)$ is AC at every compact interval of its domain $\mathcal{R}$, we say that 
the function is locally AC.
\end{defn}
The property below is essential in the following sections.
\begin{property}
 A product of two locally AC scalar functions is also locally AC.
Furthermore, the product rule also holds as shown below:
 \begin{align*}
 \nabla_\lat \left( h(\lat) f(\lat) \right) = h(\lat) \nabla_\lat f(\lat) + f(\lat) \nabla_\lat h(\lat).
 \end{align*}

\end{property}

Now, we extend the definition of locally AC to a set of functions whose domain is $\mathcal{R}^n$. 
It is known as the absolute continuity on almost every straight line (ACL) \citep{leoni2017first}.
\begin{defn}
Let $\vh(\vlat):\mathcal{R}^d \mapsto \mathcal{R}^m$ be a vector function.
Given  $\vlat_{-i} \in \mathcal{R}^{d-1}$ is fixed, let's define a function $\vh_i(\cdot):=\vh(\cdot,\vlat_{-i}):\mathcal{R}\mapsto\mathcal{R}^m$.
We say the function $\vh(\cdot)$ is locally ACL if for all $i$ and almost every point $\vlat_{-i} \in \mathcal{R}^{d-1}$, $\vh_i(\cdot)$ is locally AC.
\end{defn}
A locally ACL function is a member of the Sobolev family \citep{leoni2017first}.
Obviously, the derivative $\nabla_\lat \vh(\vlat)$ exists almost everywhere if $\vh(\vlat)$ is locally ACL.
Due to \citet{royen2010real}, a locally Lipschitz-continuous function is locally ACL.
The immediate consequence is that a function
is locally ACL and continuous if it is either locally Lipschitz-continuous or continuously differentiable.

In the following sections, we assume that the regular conditions for swapping of differentiation and integration are satisfied so that the following identify holds.
\begin{align*}
\nabla_{\varpar} \Unmyexpect{q(\lat|\varpar)}\sqr{ h(\vlat) } = \Unmyexpect{q(\lat|\varpar)}\sqr{ h(\vlat) \frac{ \nabla_{\varpar} q(\vlat|\vvarpar)}{ q(\vlat|\vvarpar) } }
\end{align*}

The regular conditions are required to use the dominated convergence theorem, which allows us to 
interchange differentiation and integration. One particular condition is
\begin{align*}
 \Unmyexpect{q(\lat|\varpar)}\sqr{ \left|\left| h(\vlat) \frac{ \nabla_{\varpar} q(\vlat|\vvarpar)}{ q(\vlat|\vvarpar) } \right|\right| }<\infty.
\end{align*}

For simplicity, we assume the regular conditions hold without explicitly mentioning them.

\section{Expectation and Conditional Expectation}
By definition, an expectation can be either non-existent or non-finite. To avoid such cases, 
we say an expectation $\Unmyexpect{q(\lat)}\sqr{ \vh(\vlat) }$ is well-defined if
\begin{align*}
\Unmyexpect{q(\lat)}\sqr{ \left|\left| \vh(\vlat) \right|\right| } <\infty,
\end{align*}
where $\left|\left| \cdot\right|\right|$ denotes the Euclidean norm.
Due to Fubini's theorem, the following identity holds in a product measure when the expectation is well-defined.
\begin{align*}
  \Unmyexpect{q(\lat_{-i})} \sqr{ \Unmyexpect{q(\lat_{i}|\lat_{-i})} \sqr{ \vh(\vlat) } } & = \Unmyexpect{q(\lat)}\sqr{ \vh(\vlat) } 
\end{align*} 
The above expression implies that conditional expectation $\Unmyexpect{q(\lat_{i}|\lat_{-i})} \sqr{ \vh(\vlat) }$ is also well-defined for almost every $\vlat_{-i}$.

For simplicity, we implicitly assume expectations are well-defined in a product measure in the paper.
For more details on Gaussian cases, please see Appendix 
\ref{app:gauss_bound_property_proofs}.

\section{Identities for Gaussian Distributions}

Now, we describe the univariate case of Stein's lemma.
\begin{lemma}
\label{lemma:stein_gauss_mean_1d}
{\bf (Stein's Lemma):}
Let $h(\cdot):\mathcal{R}\mapsto \mathcal{R}$ be locally AC.
$q(\lat)$ is an univariate Gaussian distribution denoted by $\gauss(\lat|\mu,\sigma)$, where $\mu$ is its mean and $\sigma$ is its variance.
The following first-order identity holds.
\begin{align*}
 \Unmyexpect{q} \sqr{ \frac{-\nabla_\lat q(\lat) }{q(\lat)} h(\lat) } =& \Unmyexpect{q}\sqr{ \nabla_{\lat} h(\lat)}.
\end{align*} where $\frac{-\nabla_\lat q(\lat) }{q(\lat)}= \sigma^{-1} \left( \lat-\mu \right) $.
\end{lemma}
The proof of this lemma is given at Appendix \ref{app:stein_lemma_1d_proof}.

Let's consider Bonnet's theorem, which is given below. This theorem establishes the connection between this lemma and the reparameterizable gradient for the mean $\mu$.
\begin{theorem}
\label{claim:bonnet_gauss_lemma_1d}  
{\bf (Bonnet's Theorem) :}
Let $h(\cdot):\mathcal{R} \mapsto \mathcal{R}$ be locally AC.
$q(\lat)$ is a univariate Gaussian distribution with mean $\mu$ and variance $\sigma$.
We have the following gradient identity.
\begin{align*}
\nabla_{\mu} \Unmyexpect{q} \sqr{ h(\lat) } = \Unmyexpect{q}\sqr{ \nabla_{\lat} h(\lat)}  =  \Unmyexpect{q} \sqr{ \sigma^{-1} \left( \lat-\mu \right) h(\lat) }
\end{align*}
\end{theorem}
 The proof of Bonnet's Theorem is fairly simple. First, we swap differentiation and integration.  We obtain the following expression $\nabla_{\mu} \Unmyexpect{q} \sqr{ h(\lat) } =  \Unmyexpect{q} \sqr{ \sigma^{-1} \left( \lat-\mu \right) h(\lat) }$.
By applying Lemma \ref{lemma:stein_gauss_mean_1d} to $h(\lat)$, we obtain the identity $ \Unmyexpect{q}\sqr{ \nabla_{\lat} h(\lat)}  =  \Unmyexpect{q} \sqr{ \sigma^{-1} \left( \lat-\mu \right) h(\lat) }$. 
Clearly, the re-parameterizable gradient for the mean $\mu$ is directly derived from Stein's lemma.
Now, we discuss the reparameterizable gradient for the variance.
First, we give the following lemma.
\begin{lemma}
\label{lemma:stein_gauss_var_reparam_1d}
Let $h(\cdot):\mathcal{R} \mapsto \mathcal{R}$ be locally AC.
We assume $\Unmyexpect{q}\sqr{ h(\lat)}$ is well-defined.
The following identity holds.
\begin{align*}
 \Unmyexpect{q} \sqr{ \sigma^{-2}  \left( \left( \lat-\mu \right)^2  -\sigma \right)  h(\lat) } 
=  \Unmyexpect{q}\sqr{  \sigma^{-1} \left( \lat-\mu \right)    \nabla_{\lat} h(\lat) } 
\end{align*}
\end{lemma}

The key idea of the proof is that we define an auxiliary function $f(\lat):=\sigma^{-1} \left( \lat-\mu \right) h(\lat)$ and apply Lemma \ref{lemma:stein_gauss_mean_1d} to $f(\lat)$.
By the lemma,  we have the following result: 
\begin{align*}
& \Unmyexpect{q} \sqr{ \sigma^{-2} \left( \lat-\mu \right)^2 h(\lat) } \\
=& \Unmyexpect{q} \sqr{ \sigma^{-1} \left( \lat-\mu \right) f(\lat) } \\
=& \Unmyexpect{q}\sqr{ \nabla_{\lat} f(\lat)} \\
=& \Unmyexpect{q}\sqr{ \nabla_{\lat} \left( \sigma^{-1} \left( \lat-\mu \right) h(\lat) \right) }\\
=& \Unmyexpect{q}\sqr{  \sigma^{-1} h(\lat) + \sigma^{-1} \left( \lat-\mu \right) \nabla_\lat h(\lat)  }
\end{align*}

From this expression, we can easily obtain the gradient identity.
Now, we discuss the identity to compute
the reparameterizable gradient for the variance.
\begin{lemma}
\label{claim:reparam_var_gauss_lemma_1d}  
Let $h(\lat):\mathcal{R} \mapsto \mathcal{R}$ be  locally AC.
We assume the conditions of Lemma \ref{lemma:stein_gauss_var_reparam_1d} are satisfied.
The following gradient identity holds.
\begin{align*}
\nabla_{\sigma} \Unmyexpect{q} \sqr{ h(\lat) }
= \half \Unmyexpect{q}\sqr{  \sigma^{-1} \left( \lat-\mu \right)  \nabla_{\lat} h(\lat) },
\end{align*}
\end{lemma}
Likewise, the proof of this lemma is trivial.
First, we swap differentiation and integration. We  obtain the following expression $\nabla_{\sigma} \Unmyexpect{q} \sqr{ h(\lat) } =  \half \Unmyexpect{q} \sqr{ \sigma^{-2} \left( \left( \lat-\mu \right)^2 - \sigma \right) h(\lat) }$.
After that, we obtain the above  identity by lemma \ref{lemma:stein_gauss_var_reparam_1d}.
At this point, we can see that the reparameterizable gradient for Gaussian can be derived from Stein's lemma.
Furthermore, Stein's lemma empirically gives a low-variance and unbiased gradient estimator if we allow to use the second-order information.
This idea is known as Price's theorem.
Before we discuss Price's theorem, we first describe  the following lemma.
\begin{lemma}
\label{lemma:stein_gauss_var_price_1d}
Let $h(\lat):\mathcal{R} \mapsto \mathcal{R}$ be  continuously differentiable.
Additionally, its derivative $\nabla_{\lat} h(\lat):\mathcal{R} \mapsto \mathcal{R}$ is  locally AC.
$q(\lat)$ is an univariate Gaussian distribution denoted by $\gauss(\lat|\mu,\sigma)$ with mean $\mu$ and variance $\sigma$.
We have the following identity.
\begin{align*}
 \Unmyexpect{q}\sqr{  \nabla_{\lat}^2  h(\lat) }
&=  \Unmyexpect{q}\sqr{  \sigma^{-1} \left( \lat-\mu \right)    \nabla_{\lat} h(\lat) }. 
\end{align*}
\end{lemma}

The key idea of the proof that we define auxiliary functions $f(\lat):=\nabla h(\lat)$ and apply Lemma \ref{lemma:stein_gauss_mean_1d} to $f(\lat)$.

Using the above lemmas, we obtain Price's theorem as shown below.
\begin{theorem}
\label{claim:price_gauss_lemma_1d}  
{\bf (Price's Theorem):}
Let $h(\lat):\mathcal{R} \mapsto \mathcal{R}$ be  continuously differentiable and its derivative $\nabla h(\vlat)$ be locally AC.
We further assume $\Unmyexpect{q}\sqr{ h(\lat)}$ is well-defined.
The following second-order identity holds.
\begin{align*}
\nabla_{\sigma} \Unmyexpect{q} \sqr{ h(\lat) }
= \half \Unmyexpect{q}\sqr{  \sigma^{-1} \left( \lat-\mu \right) \nabla_{\lat}^T h(\lat) } 
= \half \Unmyexpect{q} \sqr{ \nabla_{\lat}^2 h(\lat) }
\end{align*}
\end{theorem}

The above theorem can be readily shown by Lemma \ref{lemma:stein_gauss_var_price_1d}
and Lemma \ref{claim:reparam_var_gauss_lemma_1d}.
 
Now, we describe Stein's lemma for a multivariate Gaussian with arbitrary covariance structure.

\begin{lemma}
{\bf (Stein's Lemma):}
\label{lemma:stein_gauss_mean_nd}
Let $h(\vlat):\mathcal{R}^d \mapsto \mathcal{R}$ be locally ACL and continuous.
$q(\vlat)$ be a multivariate Gaussian distribution denoted by $\gauss(\vlat|\vmu,\vSigma)$ with mean $\vmu$ and variance $\vSigma$.
The following identity holds.
\begin{align*}
 \Unmyexpect{q} \sqr{ \vSigma^{-1} \left( \vlat-\vmu \right) h(\vlat) } =& \Unmyexpect{q}\sqr{ \nabla_{\lat} h(\vlat)}.
\end{align*}
\end{lemma}
The proof can be found at Appendix \ref{app:stein_lemma_nd_proof}. Bonnet's theorem and Price's theorem are given below.
The proof of Bonnet's theorem and Price's theorem can be found at Appendix \ref{app:bonnet_gauss_nd_proof} and \ref{app:price_gauss_nd_proof} respectively.

\begin{theorem}
\label{claim:bonnet_gauss_lemma}  
{\bf (Bonnet's Theorem):}
Let $h(\vlat):\mathcal{R}^d \mapsto \mathcal{R}$ be locally ACL and continuous.
$q(\vlat)$ be a multivariate Gaussian distribution denoted by $\gauss(\vlat|\vmu,\vSigma)$.
The following first-order identity holds.
\begin{align*}
\nabla_{\mu} \Unmyexpect{q} \sqr{ h(\vlat) } = \Unmyexpect{q}\sqr{ \nabla_{\lat} h(\vlat)}  =  \Unmyexpect{q} \sqr{ \vSigma^{-1} \left( \vlat-\vmu \right) h(\vlat) } 
\end{align*}
\end{theorem}

\begin{theorem}
\label{claim:price_gauss_lemma}  
{\bf (Price's Theorem):}
Let $h(\vlat):\mathcal{R}^d \mapsto \mathcal{R}$ be continuously differentiable and its derivative $\nabla h(\vlat)$ be locally ACL.
Furthermore, we assume $\Unmyexpect{q}\sqr{ h(\vlat) }$ is well-defined.
The following second-order identity holds.
\begin{align*}
\nabla_{\Sigma} \Unmyexpect{q} \sqr{ h(\vlat) }
&= \half \Unmyexpect{q}\sqr{  \vSigma^{-1} \left( \vlat-\vmu \right)    \nabla_{\lat}^T h(\vlat) } \\
&= \half \Unmyexpect{q} \sqr{ \nabla_{\lat}^2 h(\vlat) }
\end{align*}
\end{theorem}

\section{Identities for Univariate Continuous Exponential-family}
We can generalize Stein's identity to a class of exponential family.
First of all, we say a function is locally AC with its domain $(l,u)$, where $-\infty \leq l  < u \leq \infty$ if 
the function is AC at every compact interval inside its domain.
We consider the following exponential-family (EF) distribution
with $\lat \in (l,u)$,  where $-\infty \leq l  < u \leq \infty$. Furthermore, we assume $q(\lat|\vvarpar)$ is locally AC w.r.t $\lat$ and differentiable w.r.t. $\vvarpar$.
In the case when $\vphi_\lat(\vvarpar)=\vvarpar$, it can be shown that $q(\lat|\vvarpar)$ is differentiable w.r.t. $\vvarpar$. 
\begin{align*}
 q(\lat|\vvarpar) = h_\lat(\lat) \exp \crl{ \myang{\vT_\lat(\lat),\vphi_\lat(\vvarpar)}  - A_\lat( \vvarpar) }
\end{align*} where $l$ and $u$ do not depend on $\vvarpar$.

We denote the CDF of $q(\lat|\vvarpar)$ by $\psi(\lat,\vvarpar):=\int_{l}^{\lat} q(t|\vvarpar) dt$.
The following assumption is  known as the boundary condition in the literature.
\begin{align*}
\lim_{\lat \downarrow   l}  h(\lat) q(\lat|\vvarpar) = 0, \,\,\, \lim_{\lat \uparrow u} h(\lat)  q(\lat|\vvarpar) = 0
\end{align*}
\begin{lemma}
\label{lemma:stein_exp_1d}  
{\bf (Stein's Lemma):}
Let $h(\cdot):(l,u) \mapsto \mathcal{R}$ be  locally AC.
If 
the boundary condition is satisfied,
the following identity holds.
\begin{align*}
 -\myexpect_{q} \sqr{ h(\lat) \frac{\nabla_{\lat}q(\lat|\vvarpar)}{q(\lat|\vvarpar)} } =& \myexpect_{q}\sqr{ \nabla_{\lat} h(\lat)}.
\end{align*}
\end{lemma}
In the Gaussian case, we can further exploit the structure of Gaussian as shown in Lemma \ref{lemma:gauss_mono_property} so that 
the boundary condition is implicitly satisfied.
For general cases, we have to explicitly assume that the boundary condition is satisfied.
The proof is exactly the same as the proof for Lemma \ref{lemma:stein_gauss_mean_1d}
as shown in \ref{app:stein_lemma_1d_proof}.

Applying Lemma \ref{lemma:stein_exp_1d} to $\tilde{f}_i(\lat):=h(\lat)f_i(\lat)$, we obtain the implicit reparameterization trick.
\begin{theorem}
\label{claim:impl_repm_trick_1d}
{\bf (Implicit Reparametrization Trick) :}
Let $h(\cdot):(l,u) \mapsto \mathcal{R}$ be a locally AC function.
We define $f_i(\lat):=\frac{\nabla_{\varpar_{i}} \psi(\lat,\vvarpar)}{q(\lat|\vvarpar)}$ where $\varpar_i$ is a scalar.
If the conditions of Lemma \ref{lemma:stein_exp_1d} for $\tilde{f_i}(\lat)$ are satisfied, we have the following identity. 
\begin{align*}
\nabla_{\varpar_{i}} \myexpect_{q}\sqr{  h(\lat)} =  -\myexpect_{q} \sqr{ f_i(\lat) \nabla_{\lat} h(\lat)  },
\end{align*} 
\end{theorem}

\section{Identities for Exponential-family Mixtures}
\subsection{Identities for Gaussian Variance-mean Mixtures}
We consider the following Gaussian  mixture.
\begin{align*}
q(\mix,\vlat|\vmu,\valpha,\vSigma):= &\gauss(\vlat|\vmu + u(\mix)\valpha, v(\mix)\vSigma) q(\mix) \\
 q(\vlat|\vmu,\valpha,\vSigma) :=& \int q(\mix,\vlat|\vmu,\valpha,\vSigma)  d\mix 
\end{align*} where $v(\mix)>0$.

\begin{theorem}
\label{claim:bonnet_gauss_mixture}  
{\bf (Bonnet's Theorem):}
Let $h(\vlat):\mathcal{R}^d \mapsto \mathcal{R}$ be  locally ACL and continuous.
 $q(\vlat)$ is a Gaussian variance-mean mixture and $q(\mix,\vlat)$ is the joint distribution.
The following gradient identity holds.
\begin{align*}
\nabla_{\mu} \Unmyexpect{q(\lat)} \sqr{ h(\vlat) } =&  \Unmyexpect{q(\lat)}\sqr{ \nabla_{\lat} h(\vlat)} \\
\nabla_{\alpha} \Unmyexpect{q(\lat)} \sqr{ h(\vlat) } =&  \Unmyexpect{q(\mix,\lat)}\sqr{ u(\mix) \nabla_{\lat} h(\vlat)}  
\end{align*}
\end{theorem}

\begin{corollary}
If $u(\mix)$ has the following property, 
\begin{align*}
\int u(\mix) q(\mix,\vlat|\vmu,\valpha,\vSigma) d\mix  = \sum_j^k u_j(\vlat,\vmu,\valpha,\vSigma) \hat{q_j}(\vlat),
\end{align*}  where each $\hat{q}_j(\vlat)$ is a normalized distribution and $k$ is finite,
the following identity also holds.
\begin{align*}
\nabla_{\alpha} \Unmyexpect{q(\lat)} \sqr{ h(\vlat) } =\sum_{j}^k  \Unmyexpect{\hat{q_j}(\lat)}\sqr{ u_j(\vlat,\vmu,\valpha,\vSigma) \nabla_{\lat} h(\vlat)}  
\end{align*}
\end{corollary}

\begin{example}
\label{emp:skew_gauss}
A concrete example is the multivariate skew Gaussian distribution,  which can be found at Appendix
\ref{app:emp_skew_gauss}.
\end{example}

\begin{example}
\label{emp:exp_gauss}
Another example is the multivariate exponentially modified Gaussian distribution, which can be found at Appendix
\ref{app:emp_exp_gauss}.
\end{example}

\begin{theorem}
\label{claim:price_gauss_mixture}  
{\bf (Price's Theorem):}
Let $h(\vlat):\mathcal{R}^d \mapsto \mathcal{R}$ be  continuously differentiable  that its derivative $\nabla h(\vlat)$ be locally ACL.
If $\Unmyexpect{q(\mix,\lat)}\sqr{ v(\mix) h(\vlat) }$ is well-defined, 
the following  identity holds.
\begin{align*}
\nabla_{\Sigma} \Unmyexpect{q(\lat)} \sqr{ h(\vlat) }
&= \half \Unmyexpect{q(\mix,\lat)}\sqr{ v(\mix) \nabla_{\lat}^2  h(\vlat) } \\
&= \half \Unmyexpect{q(\mix,\lat)}\sqr{  \vSigma^{-1} \left( \vlat-\vmu-u(\mix)\valpha \right)    \nabla_{\lat}^T h(\vlat) } 
\end{align*}
\end{theorem}

\begin{corollary}
If $v(\mix)$ has the following property, 
\begin{align*}
\int v(\mix) q(\mix,\vlat|\vmu,\valpha,\vSigma) d\mix  = \sum_j^k v_j(\vlat,\vmu,\valpha,\vSigma) \hat{q_j}(\vlat),
\end{align*}  where each $\hat{q}_j(\vlat)$ is a normalized distribution and $k$ is finite,
the following identity also holds.
\begin{align*}
 \nabla_{\Sigma} \Unmyexpect{q(\lat)} \sqr{  h(\vlat) }  
&=  \half \Unmyexpect{q(\mix,\lat)}\sqr{ v(\mix) \nabla_{\lat}^2  h(\vlat) }\\
&= \half \sum_{j}^k  \Unmyexpect{\hat{q_j}(\lat)}\sqr{ v_j(\vlat,\vmu,\valpha,\vSigma) \nabla_{\lat}^2 h(\vlat)}  
\end{align*}
\end{corollary}

\begin{example}
\label{emp:stu_t}
A concrete example is the multivariate Student's t-distribution,
 which can be found at Appendix
\ref{app:emp_stu_t}.
\end{example}

\begin{example}
\label{emp:ninv_gauss}
Another example is the multivariate normal inverse-Gaussian distribution,
 which can be found at Appendix
\ref{app:emp_ninv_gauss}.
\end{example}

\subsection{Identities for Continuous Exponential-family Mixtures}
We consider the following EF mixtures
in a product space $\vlat=(\lat_1,\lat_2) \in (l_1,u_1) \times (l_2,u_2)$, where $-\infty \leq l_1 < u_1 \leq \infty$ and $-\infty \leq  l_2<  u_2 \leq  \infty$.
Moreover, we assume $q(\lat_1)$ $q(\lat_2|\lat_1)$, and $q(\lat_1,\lat_2)$ are locally AC, locally ACL, and continuous, respectively.
\begin{align*}
 q(\lat_1|\vvarpar) & = h_1(\lat_1) \exp \crl{ \myang{\vT_1(\lat_1),\vphi_1(\vvarpar)}  - A_1( \vvarpar) } \\
 q(\lat_2|\lat_1, \vvarpar) & = h_2(\lat_2,\lat_1) \exp \crl{ \myang{\vT_2(\lat_2,\lat_1),\vphi_2(\vvarpar)}  - A_2( \vvarpar, \lat_1) } 
\end{align*}
Let's denote the CDF of $q(\lat_1|\vvarpar)$ and the conditional CDF of $q(\lat_2|\lat_1,\vvarpar)$ by
\begin{align*}
\psi_1(\lat_1,\vvarpar)&=\int_{l_1}^{\lat_1} q(t_1|\vvarpar) dt_1 \\
\psi_2(\lat_1,\lat_2,\vvarpar)&=\int_{l_2}^{\lat_2} q(t_2|\lat_1,\vvarpar) dt_2.
\end{align*}
We define the following functions:
\begin{align*}
\vPsi(\vlat,\vvarpar) &:=\sqr{ \psi_1(\lat_1,\vvarpar), \psi_2(\lat_1,\lat_2,\vvarpar)}^T \\
\nabla_{\varpar_i}\vPsi(\vlat,\vvarpar) & :=\sqr{ \nabla_{\varpar_i} \psi_1(\lat_1,\vvarpar), \nabla_{\varpar_i}\psi_2(\lat_1,\lat_2,\vvarpar)}^T\\
\nabla_{\lat}\vPsi(\vlat,\vvarpar) &:= \begin{bmatrix}
q(\lat_1|\vvarpar) \, \,& 0 \\
\nabla_{\lat_1}\psi_2(\lat_1,\lat_2,\vvarpar)  \,  \, & q(\lat_2|\lat_1,\vvarpar)
                                    \end{bmatrix}.
\end{align*}  Note that 
$\nabla_{\lat}\vPsi(\vlat,\vvarpar)$ is a lower-triangular matrix, which is a key structure to reduce the computation cost of the implicit reparametrization trick.
Thus, it is easy to compute $\big[\nabla_{\lat}\vPsi(\vlat,\vvarpar)\big]^{-1}$.
Applying Lemma \ref{lemma:stein_exp_1d} to $\tilde{f}_{i,j}(\lat_j)$ defined below, we obtain the following identity.

\begin{theorem}
\label{claim:impl_repm_trick_2d}
{\bf (Bivariate Implicit Reparametrization Trick):}
Let $h(\cdot):(l_1,u_1) \times (l_2,u_2) \mapsto \mathcal{R}$ be locally ACL and continuous.
First, we define function $f_{i,j}(\vlat) := \ve_j^T \sqr{ \nabla_{\lat} \vPsi(\vlat,\vvarpar) }^{-1} \nabla_{\varpar_i}\vPsi(\vlat,\vvarpar)$ and 
function $\tilde{f}_{i,j}(\lat_j):= f_{i,j}(\lat_j,\vlat_{-j})  h(\lat_j,\vlat_{-j}) \prod_{k\geq j+1} q(\lat_k| \vlat_{1:(k-1)},\vvarpar ) $.
If 
the conditions of Lemma \ref{lemma:stein_exp_1d} for each $\tilde{f}_{i,j}(\lat_j)$ are satisfied,
 we have the following identity.
\begin{align*}
\nabla_{\varpar_{i}} \myexpect_{q} \big[  h(\vlat) \big] =  -\myexpect_{q} \big[  \sum_{j} f_{i,j}(\vlat) \nabla_{\lat_j} h(\vlat) \big]
\end{align*}
\end{theorem}
The proof of this theorem can be found at Appendix
\ref{app:impl_repm_trick_2d_proof}.
The identity can be easily extended to multivariate version for the implicit reparametrization trick.
\citet{figurnov2018implicit} assume that $h(\vlat)$ is continuously differentiable.
As shown in Theorem \ref{claim:impl_repm_trick_2d}, the identity holds even when $h(\vlat)$ is not continuously differentiable.

\bibliography{refs.bib}
\bibliographystyle{icml2019}


\appendix
\onecolumn
\begin{appendix}

   \section{Gradient Identities for Gaussian Distribution}
   \label{app:gauss_dist}

For completeness, we give a proof of integration by parts for AC functions, which is essential in this paper.

\begin{lemma}
\label{lemma:int_by_parts_1d}
{\bf (Integration by parts): }
Let $h(\cdot),q(\cdot):[a,b] \mapsto \mathcal{R}$ be AC functions, where $-\infty<a<b<\infty$. The following identity holds.
\begin{align*}
 h(b) q(b) - h(a) q(a)    = \int_{a}^{b} q(\lat) \nabla_\lat h(\lat) d\lat  + \int_{a}^b h(\lat) \nabla_\lat q(\lat) d\lat
\end{align*}
\end{lemma}
\begin{proof}
Since $h(\lat)$ and $q(\lat)$ are AC in $[a,b]$, we know that the product $h(\lat) q(\lat)$ is AC in $[a,b]$.
By the product rule for AC functions, the following identity holds almost everywhere for $\lat \in [a,b]$.
\begin{align}
\nabla_\lat \left( h(\lat) q(\lat) \right)  =  q(\lat) \nabla_\lat h(\lat)  + h(\lat) \nabla_\lat q(\lat) \label{eq:ing_by_parts}
\end{align}
Since $q(\lat)$ is continuous and  $h(\lat)$ is AC in $[a,b]$, we know that $q(\lat) \nabla_\lat h(\lat)$ is integrable over $[a,b]$.
Similarly, we can show $h(\lat) \nabla_\lat q(\lat)$ is integrable over $[a,b]$.
Integrating both sides of  \eqref{eq:ing_by_parts} over $[a,b]$, we obtain the identity.
\begin{align*}
 h(b) q(b) - h(a) q(a)    = \int_{a}^{b} q(\lat) \nabla_\lat h(\lat) d\lat  + \int_{a}^b h(\lat) \nabla_\lat q(\lat) d\lat
\end{align*}
\end{proof}

An alternative proof of Lemma  \ref{lemma:int_by_parts_1d} using Fubini's theorem can be found at Theorem 6 of \citet{Border-notes}.
Note that 
the condition of Fubini's theorem shown below is satisfied.
\begin{align*}
\int_{a}^{b} \int_{a}^{b}  \left| \nabla_x h(x) \nabla_y q(y) \right| dx dy < \infty,
\end{align*}
since by the definition of AC, we have
\begin{align*}
\int_{a}^{b}  \left| \nabla_x h(x)  \right| dx < \infty, \,\quad 
\int_{a}^{b}  \left| \nabla_y q(y)  \right| dy < \infty.
\end{align*}

To use integration by parts in the proof of Lemma \ref{lemma:stein_gauss_mean_1d}, we first prove the following lemma.
\begin{lemma}
\label{lemma:gauss_mono_property}
Let $h(\cdot):\mathcal{R}\mapsto \mathcal{R}$ be a locally AC function and
$q(\lat):=\gauss(\lat|\mu,\sigma)$ be an univariate Gaussian distribution with mean $\mu$ and variance $\sigma$.
If $\Unmyexpect{q}\sqr{ \nabla_\lat h(\lat) }$ is well-defined ($\Unmyexpect{q}\sqr{\left| \nabla_\lat h(\lat) \right|}<\infty$), then the following boundary conditions are satisfied:
\begin{align*}
 \lim_{\lat  \uparrow  \infty} h(\lat) q(\lat) & =0 \\
 \lim_{\lat \downarrow -\infty}  h(\lat) q(\lat) &=0 
\end{align*}
\end{lemma}

\begin{proof}
We prove that $\lim_{\lat  \uparrow  \infty} h(\lat) q(\lat)=0$ by showing that $\lim_{\lat \uparrow \infty} \inf  \left| h(\lat) \right| q(\lat) =  \lim_{\lat \uparrow \infty} \sup  \left| h(\lat) \right| q(\lat)=0$.
Given a compact interval, since $h(\lat)$ is AC, by Theorem 3.1 of \citet{RongQing-notes}, we  know that $\left|h(\lat)\right|$ is also AC  and the following identity holds almost everywhere for $t$ in the interval.
\begin{align}
-\left|\nabla_t  h(t) \right|\leq \nabla_t \left| h(t) \right| \leq \left|\nabla_t  h(t) \right| \label{ieq:abs_ac_grad}.
\end{align}
Since $\left| h(\lat) \right|$ is AC in the interval, given $c$ in the interval, by the fundamental theorem of calculus,  we have
\begin{align}
 \left| h(\lat) \right| = \left| h(c) \right| + \int_{c}^{\lat} \nabla_t \left| h(t) \right|  dt \label{eq:fund_ac_cal}.
\end{align}
Given any $ \mu \leq c  \leq \lat$, by \eqref{eq:fund_ac_cal} and then \eqref{ieq:abs_ac_grad}, we  have
\begin{align*}
\left| h(\lat) \right| q(\lat)  &= q(\lat) \left( \left| h(c) \right| +  \int_{c}^{\lat} \nabla_t \left| h(t) \right| dt \right) \\
& \leq q(\lat) \left( \left| h(c) \right| +  \int_{c}^{\lat}  \left|\nabla_t h(t) \right| dt \right) \\
& \leq  q(\lat) \left| h(c) \right| + \int_{c}^{\lat} q(t) \left|\nabla_t  h(t) \right| dt 
\end{align*} where we use the monotonicity of Gaussian: $q(\lat) \leq q(t)$ when $\mu \leq c \leq t \leq \lat$.

Therefore, by the assumption $\Unmyexpect{q(t)}\sqr{ \left| \nabla_t h(t) \right| }<\infty$, we have
\begin{align*}
 \lim_{\lat \uparrow \infty}  \sup  \left| h(\lat) \right| q(\lat)  \leq \underbrace{ \lim_{\lat \uparrow \infty}q(\lat) \overbrace{ \left|h(c)\right|}^{\text{constant}} }_{0} + \underbrace{ \int_{c}^{\infty} q(t) \left| \nabla_t h(t) \right| dt }_{<\infty} = \int_{c}^{\infty} q(t) \left| \nabla_t h(t) \right| dt,
\end{align*}
where we use the Gaussian identity that $\lim_{\lat \uparrow \infty} q(\lat)=0$.

Taking $c$ to the positive infinity, we obtain  the following result, which implies that $\lim_{\lat \uparrow \infty} \left| h(\lat) \right| q(\lat)=0$.
\begin{align*}
0 \leq \lim_{\lat \uparrow \infty} \inf  \left| h(\lat) \right| q(\lat) \leq  \lim_{\lat \uparrow \infty} \sup  \left| h(\lat) \right| q(\lat) \leq \underbrace{ \lim_{c\uparrow \infty}\int_{c}^{\infty} q(t) \left| \nabla_t h(t) \right| dt}_{0}
\end{align*}
Likewise, we can show that $\lim_{\lat \downarrow -\infty}  h(\lat)  q(\lat)=0$. 
\end{proof}

A stronger statement of 
Lemma \ref{lemma:gauss_mono_property}
can be found at Appendix
\ref{app:gauss_bound_property_proofs}.

\subsection{Proof of Lemma \ref{lemma:stein_gauss_mean_1d} and Lemma \ref{lemma:stein_exp_1d}}
\begin{proof}
\label{app:stein_lemma_1d_proof}

First, we denote the support by $(l,u)$. In the Gaussian case, $l=-\infty$ and $u=\infty$.
Since $\Unmyexpect{q}\sqr{ \nabla_{\lat} h(\lat)} $ is well-defined, we 
use the following expression to prove the claim.
\begin{align*}
 \Unmyexpect{q}\sqr{ \nabla_{\lat} h(\lat)} =
\lim_{r_1 \downarrow  l} \int_{r_1}^{c} q(\lat) \nabla_{\lat} h(\lat) d \lat
+ \lim_{r_2 \uparrow  u} \int_{c}^{r_2} q(\lat) \nabla_{\lat} h(\lat) d \lat
\end{align*} where $c \in (l,u)$ is a constant number.

Given any compact interval $[r_1,c]$, we know that $h(\lat)$ and $q(\lat)$ are AC in this interval. By integration by parts (Lemma  \ref{lemma:int_by_parts_1d}), we have
\begin{align*}
 h(c) q(c) - h(r_1) q(r_1)    = \int_{r_1}^{c} q(\lat) \nabla_\lat h(\lat) d\lat  + \int_{r_1}^c h(\lat) \nabla_\lat q(\lat) d\lat
\end{align*}
In the Gaussian case, we have $\lim_{r_1 \downarrow  l}h(r_1) q(r_1)=0$ due to Lemma \ref{lemma:gauss_mono_property}.  In non-Gaussian cases,
$\lim_{r_1 \downarrow  l}h(r_1) q(r_1)=0$ due to the boundary assumption.
Taking $r_1$ to $l$, we have
\begin{align}
 h(c) q(c) 
& = \lim_{r_1 \downarrow  l}\sqr{\int_{r_1}^{c} q(\lat) \nabla_\lat h(\lat) d\lat  + \int_{r_1}^c h(\lat) \nabla_\lat q(\lat) d\lat} \nonumber \\
& = \lim_{r_1 \downarrow  l}\int_{r_1}^{c} q(\lat) \nabla_\lat h(\lat) d\lat  + \lim_{r_1 \downarrow  l}\int_{r_1}^c h(\lat) \nabla_\lat q(\lat) d\lat  \label{eq:gauss_1d_left}.
\end{align}
Note that $\lim_{r_1 \downarrow  l}\int_{r_1}^{c} q(\lat) \nabla_\lat h(\lat) d\lat$ exists since $\Unmyexpect{q}\sqr{\left| \nabla_\lat h(\lat) \right|}<\infty$ ($\Unmyexpect{q}\sqr{\nabla_\lat h(\lat)}$ is well-defined).
Since $h(c) q(c)$ is finite, we know that $\lim_{r_1 \downarrow  l}\int_{r_1}^c h(\lat) \nabla_\lat q(\lat) d\lat$ is also finite.
Therefore, \eqref{eq:gauss_1d_left} is valid.

Likewise, given any compact interval $[c,r_2]$, by integration by parts and taking $r_2$ to $u$, we have
\begin{align}
 -h(c) q(c)   = \lim_{r_2 \uparrow  u}  \int_{c}^{r_2} q(\lat) \nabla_\lat h(\lat) d\lat  +  \lim_{r_2 \uparrow  u} \int_{c}^{r_2} h(\lat) \nabla_\lat q(\lat) d\lat  \label{eq:gauss_1d_right} .
\end{align} where $\lim_{r_2 \uparrow  u}h(r_2) q(r_2)=0$.

By \eqref{eq:gauss_1d_left} and \eqref{eq:gauss_1d_right},  we have
\begin{align*}
 \Unmyexpect{q}\sqr{ \nabla_{\lat} h(\lat)} =   - \Unmyexpect{q}\sqr{ h(\lat) \frac{\nabla_\lat q(\lat)}{q(\lat)} }, 
\end{align*} 
In the Gaussian case, we have $\nabla_\lat q(\lat)=\sigma^{-1}(\mu-\lat)q(\lat)$, which shows that the above expression is the identity.
\end{proof}

\subsection{Proof of Lemma \ref{lemma:stein_gauss_mean_nd}}
\label{app:stein_lemma_nd_proof}

\begin{proof}
We denote  the $i$-th element of $\vlat$ by $\lat_i$.
Given $\vlat_{-i}$ is known, we define a function $h_i(\lat_i):=h(\lat_i,\vlat_{-i})$.
Since  $\Unmyexpect{q}\sqr{ \nabla_{\lat} h(\vlat)}$ is well-defined, we have 
 \begin{align*}
  \Unmyexpect{q}\sqr{ \nabla_{\lat_i} h(\vlat)} 
=  \Unmyexpect{q(\lat_{-i})q(\lat_i|\lat_{-i})}\sqr{ \nabla_{\lat_i} h(\lat_i, \vlat_{-i})}  
=  \Unmyexpect{q(\lat_{-i})}\sqr{ \Unmyexpect{q(\lat_i|\lat_{-i})} \sqr{ \nabla_{\lat_i} h_i(\lat_i)} } 
 \end{align*}
 
Without loss of generality, we assume $\lat_i$ is the last element of $\vlat$. It is possible since we can permute the elements of $\vlat$ to achieve that.
Therefore, we can re-express the mean and the covariance matrix as below.
\begin{align*}
 \vmu  = \begin{bmatrix}
         \vmu_{-i} \\
         \mu_i
        \end{bmatrix}, \,\quad
 \vSigma = \begin{bmatrix}
            \vSigma_{-i,-i} & \vSigma_{-i,i} \\
            \vSigma_{-i,i}^T & \Sigma_{i,i}
            \end{bmatrix}
\end{align*}

Note that $q(\lat_i|\vlat_{-i})$ is an univariate Gaussian distribution denote by $\gauss(\lat_i|m,\sigma)$, where
\begin{align}
m & = \mu_i + \vSigma_{-i,i}^T \vSigma_{-i,-i}^{-1}\left( \vlat_{-i} - \vmu_{-i} \right) \label{eq:cond_gauss_mean} \\
\sigma &= \Sigma_{i,i} - \vSigma_{-i,i}^T \vSigma_{-i,-i}^{-1}\vSigma_{-i,i} \label{eq:cond_gauss_var}.
\end{align}

Since 
$ h_i(\lat_i)$ is locally AC, 
we have the following result by applying Lemma \ref{lemma:stein_gauss_mean_1d} to $h_i(\lat_i)$.
\begin{align*}
 \Unmyexpect{q} \sqr{ \nabla_{\lat_i} h(\vlat)} 
&= \Unmyexpect{q(\lat_{-i})}\sqr{ \Unmyexpect{q(\lat_i|\lat_{-i})} \sqr{ \nabla_{\lat_i} h_i(\lat_i)} }  \\
&= \Unmyexpect{q(\lat_{-i})} \sqr{ \Unmyexpect{q(\lat_i|\lat_{-i})} \sqr{ \sigma^{-1} \left( \lat_i - m \right) h_i(\lat_i) } }\\
&= \Unmyexpect{q} \sqr{ \sigma^{-1} \left( \lat_i - m \right) h_i(\lat_i) } \\
&= \Unmyexpect{q} \sqr{ \sigma^{-1} \left( \lat_i - m \right) h(\vlat) } 
\end{align*}
Recall that by assumptions the above expectations are well-defined. 
It can be verified that
\begin{align}
\sigma^{-1} \left( \lat_i - m \right)  &= \ve_i^T \vSigma^{-1} \left( \vlat-\vmu \right) \label{eq:gauss_cond_exp_res}
\end{align}
where $\ve_i$ is an one-hot vector where it has all zero elements except the $i$-th element with value $1$.

Using the result at \eqref{eq:gauss_cond_exp_res}, we have
\begin{align}
 \Unmyexpect{q} \sqr{ \nabla_{\lat_i} h(\vlat)} 
= \Unmyexpect{q} \sqr{ \sigma^{-1} \left( \lat_i - m \right) h(\vlat) } 
= \Unmyexpect{q} \sqr{ \ve_i^T \vSigma^{-1} \left( \vlat-\vmu \right) h(\vlat) } \label{eq:nd_stein_lemma_gauss}
\end{align}

Therefore, we have
\begin{align*}
 \Unmyexpect{q}\sqr{  \nabla_{\lat} h(\vlat) }  
&= \Unmyexpect{q} \sqr{  \vSigma^{-1} \left( \vlat-\vmu \right) h(\vlat) } 
\end{align*}

\end{proof}

\subsection{Proof of Theorem \ref{claim:bonnet_gauss_lemma}}
\label{app:bonnet_gauss_nd_proof}
\begin{proof}
We swap integration and differentiation and obtain the following result.
\begin{align*}
 \nabla_{\mu} \Unmyexpect{q} \sqr{ h(\vlat) } &= \int h(\vlat) \nabla_{\mu} \gauss(\vlat|\vmu,\vSigma) d\vlat \\
 &= \int h(\vlat) \vSigma^{-1} \left( \vlat-\vmu \right) \gauss(\vlat|\vmu,\vSigma) d\vlat \\
 &= \Unmyexpect{q}\sqr{ \vSigma^{-1} \left( \vlat-\vmu \right) h(\vlat) } 
\end{align*} which is known as the score function estimator.


Using Lemma \ref{lemma:stein_gauss_mean_nd} to move from line 1 to line 2, we obtain the gradient identity given below.
\begin{align*}
 \nabla_{\mu} \Unmyexpect{q} \sqr{ h(\vlat) }
 &= \Unmyexpect{q}\sqr{ \vSigma^{-1} \left( \vlat-\vmu \right) h(\vlat) } \\
&= \Unmyexpect{q}\sqr{  \nabla_{\lat} h(\vlat) }  
\end{align*} which is known as the re-parametrization trick for the mean $\vmu$.

\end{proof}

\subsection{Proof of Theorem \ref{claim:price_gauss_lemma}}
\label{app:price_gauss_nd_proof}
To prove Theorem \ref{claim:price_gauss_lemma}, we first prove the following lemma, which is a multivariate extension of Lemma \ref{lemma:stein_gauss_var_reparam_1d}.
By convention, $\ve_i$ is an one-hot vector where it has all zero elements except the $i$-th element with value $1$.
\begin{lemma}
\label{lemma:stein_gauss_var_reparam}
Let $h(\vlat):\mathcal{R}^d \mapsto \mathcal{R}$ be  locally ACL and continuous.
We  define an auxiliary vector function $\vf(\vlat)=  \vSigma^{-1} \left( \vlat-\vmu \right)   h(\vlat)  $.
If $\Unmyexpect{q}\sqr{h(\vlat) }$ is well-defined, 
the following identity holds.
\begin{align*}
 \Unmyexpect{q} \sqr{ \vSigma^{-1}  \sqr{\left( \vlat-\vmu \right)  \left( \vlat-\vmu \right)^T -\vSigma } \vSigma^{-1}  h(\vlat) } 
&=  \Unmyexpect{q}\sqr{  \vSigma^{-1} \left( \vlat-\vmu \right)    \nabla_{\lat}^T h(\vlat) }
\end{align*}
\end{lemma}
\begin{proof}

We define a scalar function $f_i(\vlat):=\ve_i^T\vf(\vlat)$.
By applying Lemma 
\ref{lemma:stein_gauss_mean_nd} to $f_i(\vlat)$, we have
\begin{align*}
& \Unmyexpect{q} \sqr{ \vSigma^{-1} \left( \vlat-\vmu \right) \underbrace{ \left( \vlat-\vmu \right)^T \vSigma^{-1} \ve_i h(\vlat)}_{\text{scalar}} } \\
=& 
 \Unmyexpect{q} \sqr{ \vSigma^{-1} \left( \vlat-\vmu \right)  \ve_i^T \underbrace{\vSigma^{-1} \left( \vlat-\vmu \right)  h(\vlat)}_{=\vf(\vlat)} }\\
=& \Unmyexpect{q} \sqr{ \vSigma^{-1} \left( \vlat-\vmu \right) \ve_i^T \vf(\vlat) } \\
 =& \Unmyexpect{q}\sqr{   \nabla_{\lat} \left( \ve_i^T\vf(\vlat) \right) }
\end{align*}
Recall that
$
\Unmyexpect{q}\sqr{   \nabla_{\lat} \left( \ve_i^T\vf(\vlat) \right) }
= \Unmyexpect{q}\sqr{    \vSigma^{-1} \ve_i  h(\vlat) +  \ve_i^T  \vSigma^{-1}  \left( \vlat-\vmu \right)  \nabla_\lat h(\vlat) }.
$

Therefore, we know that
\begin{align*}
 \Unmyexpect{q} \sqr{ \ve_j^T \vSigma^{-1} \left( \vlat-\vmu \right) \left( \vlat-\vmu \right)^T \vSigma^{-1} \ve_i h(\vlat)  } 
=& \Unmyexpect{q} \sqr{ \ve_j^T \vSigma^{-1} \left( \vlat-\vmu \right) \ve_i^T \vf(\vlat) } \\
=& \Unmyexpect{q}\sqr{  \ve_j^T \nabla_{\lat} \left( \ve_i^T\vf(\vlat) \right) } \\
=& \Unmyexpect{q}\sqr{  \ve_j^T  \vSigma^{-1} \ve_i  h(\vlat) +  \ve_i^T  \vSigma^{-1}  \left( \vlat-\vmu \right)  \ve_j^T \nabla_\lat h(\vlat) } 
\end{align*}

Since $\Unmyexpect{q}\sqr{h(\vlat)}$ is also well-defined, we have
\begin{align*}
 \Unmyexpect{q} \sqr{ \ve_i^T \vSigma^{-1} \left( \left( \vlat-\vmu \right) \left( \vlat-\vmu \right)^T -\vSigma\right) \vSigma^{-1} \ve_j h(\vlat)  } 
=& \Unmyexpect{q} \sqr{ \ve_j^T \vSigma^{-1} \left( \left( \vlat-\vmu \right) \left( \vlat-\vmu \right)^T -\vSigma\right) \vSigma^{-1} \ve_i h(\vlat)  } \\
=& \Unmyexpect{q}\sqr{   \ve_i^T  \vSigma^{-1}  \left( \vlat-\vmu \right)  \ve_j^T \nabla_\lat h(\vlat) } \\
=& \Unmyexpect{q}\sqr{   \ve_i^T  \vSigma^{-1}  \left( \vlat-\vmu \right)   \nabla_\lat^T h(\vlat) \ve_j} ,
\end{align*} which implies that
\begin{align*}
 \Unmyexpect{q} \sqr{  \vSigma^{-1} \left( \left( \vlat-\vmu \right) \left( \vlat-\vmu \right)^T -\vSigma\right) \vSigma^{-1}  h(\vlat)  } 
= \Unmyexpect{q}\sqr{    \vSigma^{-1}  \left( \vlat-\vmu \right)   \nabla_\lat^T h(\vlat) }.
\end{align*}

\end{proof}

Next, we prove the following lemma, which is a multivariate extension of Lemma \ref{claim:reparam_var_gauss_lemma_1d}.

\begin{lemma}
\label{claim:reparam_var_gauss_lemma}  
Let $h(\vlat):\mathcal{R}^d \mapsto \mathcal{R}$ be  locally ACL and continuous.
We assume the conditions of Lemma \ref{lemma:stein_gauss_var_reparam} are satisfied. 
The following gradient identity holds.
\begin{align*}
\nabla_{\Sigma} \Unmyexpect{q} \sqr{ h(\vlat) }
&= \half \Unmyexpect{q}\sqr{  \vSigma^{-1} \left( \vlat-\vmu \right)    \nabla_{\lat}^T h(\vlat) }
\end{align*}
\end{lemma}

\begin{proof}
By the assumptions, we can interchange the integration and differentiation to obtain  the following result.
\begin{align*}
 \nabla_{\Sigma} \Unmyexpect{q} \sqr{ h(\vlat) } &= \int h(\vlat) \nabla_{\Sigma} \gauss(\vlat|\vmu,\vSigma) d\vlat \\
 &= \half \int h(\vlat) \vSigma^{-1}  \sqr{\left( \vlat-\vmu \right)  \left( \vlat-\vmu \right)^T -\vSigma } \vSigma^{-1} \gauss(\vlat|\vmu,\vSigma) d\vlat \\
 &= \half \Unmyexpect{q} \sqr{ \vSigma^{-1}  \sqr{\left( \vlat-\vmu \right)  \left( \vlat-\vmu \right)^T -\vSigma } \vSigma^{-1}  h(\vlat) }
\end{align*} which is known as the score function estimator.


By Lemma \ref{lemma:stein_gauss_var_reparam}, we have
\begin{align*}
 \nabla_{\Sigma} \Unmyexpect{q} \sqr{ h(\vlat) } 
&= \half \Unmyexpect{q} \sqr{ \vSigma^{-1}  \sqr{\left( \vlat-\vmu \right)  \left( \vlat-\vmu \right)^T -\vSigma } \vSigma^{-1}  h(\vlat) } \\
&= \half \Unmyexpect{q}\sqr{  \vSigma^{-1} \left( \vlat-\vmu \right)    \nabla_{\lat}^T h(\vlat) } .
\end{align*}
\end{proof}

The following lemma is  also useful when we prove Theorem \ref{claim:price_gauss_lemma}. This lemma is a multivariate extension of Lemma \ref{lemma:stein_gauss_var_price_1d}.
\begin{lemma}
\label{lemma:stein_gauss_var_price}
Let a function $h(\vlat):\mathcal{R}^d \mapsto \mathcal{R}$ be continuously differentiable and its derivative $\nabla_{\lat} h(\vlat):\mathcal{R}^d \mapsto \mathcal{R}^d$ be locally ACL.
$q(\vlat)$ is a multivariate Gaussian distribution denoted by $\gauss(\vlat|\vmu,\vSigma)$ with mean $\vmu$ and variance $\vSigma$.
The following identity holds.
\begin{align}
 \Unmyexpect{q}\sqr{  \nabla_{\lat}^2  h(\vlat) }
&=  \Unmyexpect{q}\sqr{  \vSigma^{-1} \left( \vlat-\vmu \right)    \nabla_{\lat}^T h(\vlat) }, 
 \label{eq:asym_gauss_price_eq} 
\end{align}
\end{lemma}

\begin{proof}
We define a scalar function $g_i(\vlat)=\nabla_{\lat}^T h(\vlat)\ve_i$.
By applying Lemma 
\ref{lemma:stein_gauss_mean_nd} to $g_i(\vlat)$, we have
\begin{align*}
 \Unmyexpect{q} \sqr{ \vSigma^{-1} \left( \vlat-\vmu \right) g_i(\vlat) } =& \Unmyexpect{q}\sqr{ \nabla_{\lat} g_i(\vlat)}.
\end{align*}
Therefore, we have
\begin{align*}
 \Unmyexpect{q} \sqr{ \vSigma^{-1} \left( \vlat-\vmu \right) \nabla_{\lat}^T h(\vlat) \ve_i } 
=&  \Unmyexpect{q} \sqr{ \vSigma^{-1} \left( \vlat-\vmu \right) g_i(\vlat) } \\
=& \Unmyexpect{q}\sqr{ \nabla_{\lat} g_i(\vlat)} \\
=& \Unmyexpect{q}\sqr{ \nabla_{\lat} \left(  \nabla_{\lat}^T h(\vlat) \ve_i \right) } \\
=& \Unmyexpect{q}\sqr{  \nabla_{\lat}^2 h(\vlat) \ve_i },
\end{align*} which implies that
\begin{align}
 \Unmyexpect{q}\sqr{  \nabla_{\lat}^2  h(\vlat) }
=  \Unmyexpect{q}\sqr{  \vSigma^{-1} \left( \vlat-\vmu \right)    \nabla_{\lat}^T h(\vlat) } 
\end{align}

\end{proof}

Now, it is time for us to prove Theorem \ref{claim:price_gauss_lemma}.
\begin{proof}

 
Note that all conditions of Lemma \ref{lemma:stein_gauss_var_price} are satisfied. By Lemma \ref{lemma:stein_gauss_var_price}, we have
\begin{align}
 \Unmyexpect{q}\sqr{  \nabla_{\lat}^2  h(\vlat) }
&=  \Unmyexpect{q}\sqr{  \vSigma^{-1} \left( \vlat-\vmu \right)    \nabla_{\lat}^T h(\vlat) },
\label{eq:price_eq_gauss_p1}
\end{align} 

Since all conditions of Lemma \ref{claim:reparam_var_gauss_lemma}  are satisfied, by Lemma  \ref{claim:reparam_var_gauss_lemma}, we have
\begin{align}
\nabla_{\Sigma} \Unmyexpect{q} \sqr{ h(\vlat) }
= \half \Unmyexpect{q}\sqr{  \vSigma^{-1} \left( \vlat-\vmu \right)    \nabla_{\lat}^T h(\vlat) } 
\label{eq:price_eq_gauss_p2}
\end{align}

Therefore, by 
\eqref{eq:price_eq_gauss_p1}
and 
\eqref{eq:price_eq_gauss_p2}
we have
\begin{align*}
\nabla_{\Sigma} \Unmyexpect{q} \sqr{ h(\vlat) }
= \half \Unmyexpect{q}\sqr{  \vSigma^{-1} \left( \vlat-\vmu \right)    \nabla_{\lat}^T h(\vlat) } 
= \half \Unmyexpect{q}\sqr{  \nabla_{\lat}^2  h(\vlat) }
\end{align*}

\end{proof}

\subsection{Well-defined Expectations}
\label{app:gauss_bound_property_proofs}

\begin{lemma}
\label{lemma:gauss_bound_property}
Let $h(\cdot):\mathcal{R}\mapsto \mathcal{R}$ be locally AC and continuous.
$q(\lat):=\gauss(\lat|\mu,\sigma)$ is an univariate Gaussian distribution with mean $\mu$ and variance $\sigma$.
Let $C:=\left|  h(\mu) \right| \Unmyexpect{q}\sqr{\left| \sigma^{-1}(\lat-\mu)\right|}=\left|h(\mu) \right|\sqrt{\frac{2}{\sigma\pi}}<\infty$.
If $\Unmyexpect{q}\sqr{\left| \nabla_\lat h(\lat) \right|}<\infty$, then
\begin{align*}
\Unmyexpect{q}\sqr{\left| h(\lat) \right|} <\infty,\quad
\Unmyexpect{q}\sqr{\left| \sigma^{-1} (\lat-\mu)   h(\lat) \right|} \leq  \Unmyexpect{q}\sqr{\left| \nabla_\lat h(\lat) \right|}+ C<\infty.
\end{align*}
\end{lemma}
\begin{proof}
This proof is inspired by the note of \citet{Sourav-notes}. 
Firstly, we  assume  that 
$\Unmyexpect{q}\sqr{\left| \sigma^{-1} (\lat-\mu)   h(\lat) \right|}<\infty$.
Since $ h(\lat) $ is continuous on $S:=\{\lat| \left|\sigma^{-1} (\lat-\mu)\right|\leq 1\}$, 
by extreme value theorem, we know that
$m=\inf_{\lat \in S} h(\lat)$ and $M=\sup_{\lat \in  S} h(\lat)$ both exist and are bounded.
Consequently, when $\lat  \in S$, we have
\begin{align*}
\left| h(\lat) \right| \leq  \left| m \right| + \left| M \right|  < \infty
\end{align*}

When $\lat  \not\in S$, we know that $\left| \sigma^{-1} (\lat-\mu) \right|>1$.
Therefore, we have
\begin{align*}
\left| h(\lat) \right| \leq \left| \sigma^{-1} (\lat-\mu) h(\lat) \right| 
\end{align*}
By the above inequalities, for $\lat \in \mathcal{R}$, we have
\begin{align*}
\left| h(\lat) \right| \leq \left| \sigma^{-1} (\lat-\mu) h(\lat) \right| + \left| m \right| + \left| M \right|
\end{align*}
Therefore, we can show that 
$\Unmyexpect{q}\sqr{\left|   h(\lat) \right|}$ is bounded since
\begin{align*}
\Unmyexpect{q}\sqr{\left| h(\lat) \right|} \leq \Unmyexpect{q}\sqr{\left| \sigma^{-1} (\lat-\mu) h(\lat) \right|} + \left| m \right| + \left| M \right|<\infty
\end{align*}
Now, we show that
$\Unmyexpect{q}\sqr{\left| \sigma^{-1} (\lat-\mu)   h(\lat) \right|}\leq \Unmyexpect{q}\sqr{\left| \nabla_\lat h(\lat) \right|}+ C$.

Since
$\Unmyexpect{q}\sqr{\left| \nabla_\lat   h(\lat) \right|}<\infty$,
by the triangular inequality, we have
\begingroup
\begin{align*}
\Unmyexpect{q}\sqr{\left| \sigma^{-1} (\lat-\mu)   h(\lat) \right|} 
\leq &\Unmyexpect{q}\sqr{ \sigma^{-1} \left| \lat-\mu\right| \left(  \left| h(\mu) \right| + \left| h(\lat)-h(\mu)  \right| \right)}  \\
= &\Unmyexpect{q}\sqr{\sigma^{-1} \left| \lat-\mu\right|  \left| h(\lat)-h(\mu)  \right| } + \underbrace{ \left| h(\mu) \right| \Unmyexpect{q}\sqr{\left| \sigma^{-1}(\lat-\mu)\right|}}_{=C} 
\end{align*}
\vspace{-0.5cm}
\endgroup
Notice that
\begin{align*}
\Unmyexpect{q}\sqr{\sigma^{-1} \left| \lat-\mu\right|  \left| h(\lat)-h(\mu)  \right| } 
= & \underbrace{\int_{\mu}^{\infty} \sigma^{-1} \left( \lat-\mu \right)\left| h(\lat)-h(\mu)  \right| q(\lat)d\lat}_{\text{term I}}
+ \underbrace{ \int_{-\infty}^{\mu} \sigma^{-1} \left( \mu-\lat \right)\left| h(\lat)-h(\mu)  \right| q(\lat) d\lat}_{\text{term II}}
\end{align*} 
\vspace{-0.4cm}
Now, we show that 
$\Unmyexpect{q}\sqr{\sigma^{-1} \left| \lat-\mu\right|  \left| h(\lat)-h(\mu)  \right| } \leq \Unmyexpect{q}\sqr{\left| \nabla_\lat h(\lat) \right|}$
since both terms are bounded as shown below.
\begin{align*}
\int_{\mu}^{\infty} \sigma^{-1} \left( \lat-\mu \right)\left| h(\lat)-h(\mu)  \right| q(\lat)d\lat & \leq \int_{\mu}^{\infty} \left|\nabla_\lat h(\lat) \right| q(\lat) d\lat \\
\int_{-\infty}^{\mu} \sigma^{-1} \left( \mu-\lat \right)\left| h(\lat)-h(\mu)  \right| q(\lat) d\lat & \leq \int_{-\infty}^{\mu} \left|\nabla_\lat h(\lat) \right| q(\lat) d\lat 
\end{align*}

Since $h(\lat)$ is locally AC, by the fundamental theorem of calculus, we have
\begin{align*}
 \int_{\mu}^{\infty} \sigma^{-1} \left( \lat-\mu \right)\left| h(\lat)-h(\mu)  \right| q(\lat)d\lat 
&=  \int_{\mu}^{\infty} \sigma^{-1}\left( \lat-\mu \right)q(\lat)\left| \int_{\mu}^{\lat} \nabla_t h(t) dt \right| d\lat \\
&\leq  \int_{\mu}^{\infty} \sigma^{-1}\left( \lat-\mu \right) q(\lat) \int_{\mu}^{\lat} \left|\nabla_t h(t) \right| dt d\lat \\
&=  \int_{\mu}^{\infty} \left|\nabla_t h(t) \right| \int_{t}^{\infty} \sigma^{-1} \left( \lat-\mu \right) q(\lat)  d\lat dt
\end{align*}
If $\int_{\mu}^{\infty} \left|\nabla_t h(t) \right| \int_{t}^{\infty} \left| \sigma^{-1}\left( \lat-\mu \right) q(\lat) \right| d\lat dt<\infty$,
by Fubini's theorem, we obtain the last step in the above expression.

Now, we  show that $\int_{\mu}^{\infty} \left|\nabla_t h(t) \right| \int_{t}^{\infty} \left| \sigma^{-1}\left( \lat-\mu \right) q(\lat) \right| d\lat dt<\infty$ since
\begin{align*}
 \int_{\mu}^{\infty} \left|\nabla_t h(t) \right| \int_{t}^{\infty} \left| \sigma^{-1}\left( \lat-\mu \right) q(\lat) \right| d\lat  dt
 = & \int_{\mu}^{\infty} \left|\nabla_t h(t) \right| \int_{t}^{\infty} \sigma^{-1}\left( \lat-\mu \right) q(\lat) d\lat dt  
\end{align*}
Since $q(\lat)$ is Gaussian with mean $\mu$ and variance $\sigma$, we have
\begin{align*}
  \int_{t}^{\infty} \sigma^{-1}\left( \lat-\mu \right)  q(\lat) d\lat &= 
  \int_{t}^{\infty} \sigma^{-1}\left( \lat-\mu \right) \frac{1}{\sqrt{2\pi \sigma}} \exp\left(- \frac{(\lat-\mu)^2}{2\sigma} \right)  d\lat \\
  &= -\frac{1}{\sqrt{2\pi \sigma}}\exp\left( -\frac{(\lat-\mu)^2}{2\sigma} \right)\Big|_{t}^{\infty} \\
  &= q(t)
\end{align*}
Therefore,  we have
\begin{align*}
\int_{\mu}^{\infty} \sigma^{-1} \left( \lat-\mu \right)\left| h(\lat)-h(\mu)  \right| q(\lat)d\lat &\leq
 \int_{\mu}^{\infty} \left|\nabla_t h(t) \right| \int_{t}^{\infty} \left| \sigma^{-1}\left( \lat-\mu \right) q(\lat) \right| d\lat  dt \\
 &=  \int_{\mu}^{\infty} \left|\nabla_t h(t) \right| q(t) dt \\
 &=  \int_{\mu}^{\infty} \left|\nabla_\lat h(\lat) \right| q(\lat) d\lat
\end{align*}

Similarly, we can show
$\int_{-\infty}^{\mu} \sigma^{-1} \left( \mu-\lat \right)\left| h(\lat)-h(\mu)  \right| q(\lat) d\lat
\leq \int_{-\infty}^{\mu} \left|\nabla_\lat h(\lat) \right| q(\lat) d\lat.
$
\end{proof}

\begin{lemma}
\label{lemma:gauss_bound_property_cont}
Let $h(\cdot):\mathcal{R}\mapsto \mathcal{R}$ be continuously differentiable and
its derivative be locally AC.
$q(\lat):=\gauss(\lat|\mu,\sigma)$ is an univariate Gaussian distribution with mean $\mu$ and variance $\sigma$.
If $\Unmyexpect{q}\sqr{\left| \nabla_\lat^2 h(\lat) \right|}<\infty$, then
\begin{align*}
&\Unmyexpect{q}\sqr{\left| \nabla_\lat h(\lat) \right|}  <\infty,\quad
\Unmyexpect{q}\sqr{\left| \sigma^{-1} (\lat-\mu)  \nabla_\lat h(\lat) \right|} <\infty,\\
&\Unmyexpect{q}\sqr{\left| h(\lat) \right|} <\infty,\quad\quad\,\,
\Unmyexpect{q}\sqr{\left| \sigma^{-1} (\lat-\mu)   h(\lat) \right|} <\infty.
\end{align*}
\end{lemma}
\begin{proof}
We first define an auxiliary function $f(\lat):=\nabla_\lat h(\lat)$.
Since $\Unmyexpect{q}\sqr{\left| \nabla_\lat^2 h(\lat) \right|}=\Unmyexpect{q}\sqr{\left| \nabla_\lat f(\lat) \right|}<\infty$,
by applying
Lemma \ref{lemma:gauss_bound_property} to $f(\lat)$, we have
\begin{align*}
\Unmyexpect{q}\sqr{\left| \nabla_\lat h(\lat) \right|}  <\infty,\quad
\Unmyexpect{q}\sqr{\left| \sigma^{-1} (\lat-\mu)  \nabla_\lat h(\lat) \right|} <\infty.
\end{align*}
Furthermore,
since $
\Unmyexpect{q}\sqr{\left| \nabla_\lat h(\lat) \right|}  <\infty$,
by applying
Lemma \ref{lemma:gauss_bound_property} to $h(\lat)$, we have
\begin{align*}
\Unmyexpect{q}\sqr{\left| h(\lat) \right|} <\infty,\quad
\Unmyexpect{q}\sqr{\left| \sigma^{-1} (\lat-\mu)   h(\lat) \right|} <\infty.
\end{align*}
 
\end{proof}

\begin{lemma}
\label{lemma:gauss_bound_property_nd}
Let $h(\cdot):\mathcal{R}^d\mapsto \mathcal{R}$ be locally ACL and continuous.
$q(\vlat):=\gauss(\vlat|\vmu,\vSigma)$ is a multivariate Gaussian distribution with mean $\vmu$ and variance $\vSigma$.
Let $\left|\left|\cdot\right|\right|$ denote the Euclidean norm.
If $\Unmyexpect{q}\sqr{\left|\left| \nabla_\lat h(\vlat) \right|\right|}<\infty$ and 
$\Unmyexpect{q}\sqr{\left| h(\vlat) \right|} <\infty$, then
\begin{align*}
\Unmyexpect{q}\sqr{\left|\left| \vSigma^{-1} (\vlat-\vmu)   h(\vlat) \right|\right|} <\infty.
\end{align*}
\end{lemma}

\begin{proof}
We prove that
$\Unmyexpect{q}\sqr{\left|\left| \vSigma^{-1} (\vlat-\vmu)   h(\vlat) \right|\right|}<\infty$ by
showing
$\Unmyexpect{q}\sqr{\left|\ve_i^T \vSigma^{-1} (\vlat-\vmu)   h(\vlat) \right|}<\infty$ for any natural number $1 \leq i \leq d$.
Since  $\Unmyexpect{q}\sqr{\left|\left| \nabla_\lat h(\vlat) \right|\right|}<\infty$, we know that
\begin{align*}
\Unmyexpect{q(\lat_{-i})q(\lat_i|\lat_{-i})}\sqr{\left| \nabla_{\lat_i} h(\vlat) \right|} 
= \Unmyexpect{q}\sqr{\left| \nabla_{\lat_i} h(\vlat) \right|} < \infty.
\end{align*}
The above inequality implies that for almost every $\vlat_{-i}$, the following identity holds.
\begin{align*}
\Unmyexpect{q(\lat_i|\lat_{-i})}\sqr{\left| \nabla_{\lat_i} h(\vlat) \right|}
= \Unmyexpect{q(\lat_i|\lat_{-i})}\sqr{\left| \nabla_{\lat_i} h_i(\lat_i) \right|}
<\infty
\end{align*} where we define an auxiliary function $h_i(\lat_i):=h(\lat_i,\vlat_{-i})$.
Without loss of generality, we assume $\lat_i$ is the last element of $\vlat$. It is possible since we can permute the elements of $\vlat$ to achieve that.
Therefore, we can re-express the mean and the covariance matrix as below.
\begin{align*}
 \vmu  = \begin{bmatrix}
         \vmu_{-i} \\
         \mu_i
        \end{bmatrix}, \,\quad
 \vSigma = \begin{bmatrix}
            \vSigma_{-i,-i} & \vSigma_{-i,i} \\
            \vSigma_{-i,i}^T & \Sigma_{i,i}
            \end{bmatrix}
\end{align*}

Note that $q(\lat_i|\vlat_{-i})$ is an univariate Gaussian distribution denote by $\gauss(\lat_i|m,\sigma)$, where
\begin{align*}
m & = \mu_i + \vSigma_{-i,i}^T \vSigma_{-i,-i}^{-1}\left( \vlat_{-i} - \vmu_{-i} \right)\\ 
\sigma &= \Sigma_{i,i} - \vSigma_{-i,i}^T \vSigma_{-i,-i}^{-1}\vSigma_{-i,i} 
\end{align*}
Note that $h_i(\lat_i)$ is locally AC w.r.t. $\lat_i$ since $h(\vlat)$ is locally ACL.
Since for almost every $\vlat_{-i}$, $\Unmyexpect{q(\lat_i|\lat_{-i})}\sqr{\left| \nabla_{\lat_i} h_i(\lat_i) \right|}<\infty$, by applying Lemma \ref{lemma:gauss_bound_property} to $h_i(\lat_i)$, we have
\begin{align*}
\Unmyexpect{q(\lat_i|\lat_{-i})}\sqr{\left| \sigma^{-1} \left( \lat_i - m \right) h_i(\lat_i) \right|}
\leq \Unmyexpect{q(\lat_i|\lat_{-i})}\sqr{\left| \nabla_{\lat_i} h_i(\lat_i) \right|}+ C_i
\end{align*}
where  $C_i:=\left|  h_i(m) \right| \Unmyexpect{q(\lat_i|\lat_{-i})}\sqr{\left| \sigma^{-1}(\lat_i-m)\right|}=\left|h_i(m) \right|\sqrt{\frac{2}{\sigma\pi}}$.

Since $\Unmyexpect{q}\sqr{\left| h(\vlat) \right|} <\infty$, then
\begin{align*}
\Unmyexpect{q(\lat_{-i})}\sqr{ C_i } & = \Unmyexpect{q(\lat_{-i})}\sqr{ \left|h_i(m) \right|} \sqrt{\frac{2}{\sigma\pi}} 
\leq \underbrace{ \Unmyexpect{q(\lat_{-i})}\sqr{ \Unmyexpect{q(\lat_i)} \sqr{ \left|h_i(\lat_i) \right|} }}_{ \Unmyexpect{q}\sqr{ \left| h(\boldsymbol{\lat}) \right| } } \sqrt{\frac{2}{\sigma\pi}} <\infty
\end{align*}

Consequently,  we have
\begin{align*}
\Unmyexpect{q(\lat_{-i})q(\lat_i|\lat_{-i})}\sqr{\left| \sigma^{-1} \left( \lat_i - m \right) h_i(\lat_i) \right|}
\leq \underbrace{\Unmyexpect{q(\lat_{-i})}\sqr{ \Unmyexpect{q(\lat_i|\lat_{-i})}\sqr{\left| \nabla_{\lat_i} h_i(\lat_i) \right|}} }_{\Unmyexpect{q}\sqr{ \left| \nabla_{\lat_i} h(\boldsymbol{\lat})\right| }} + \Unmyexpect{q(\lat_{-i})}\sqr{ C_i }
<\infty
\end{align*}

It can be verified that
\begin{align*}
\sigma^{-1} \left( \lat_i - m \right)  &= \ve_i^T \vSigma^{-1} \left( \vlat-\vmu \right) 
\end{align*}
where $\ve_i$ is an one-hot vector where it has all zero elements except the $i$-th element with value $1$.

Therefore, we have
\begin{align*}
 \Unmyexpect{q}\sqr{\left| \ve_i^T \vSigma^{-1} \left( \vlat-\vmu \right) h(\vlat) \right|} 
= \Unmyexpect{q(\lat_{-i})q(\lat_i|\lat_{-i})}\sqr{\left| \sigma^{-1} \left( \lat_i - m \right) h(\vlat) \right|} 
=  \Unmyexpect{q(\lat_{-i})q(\lat_i|\lat_{-i})}\sqr{\left| \sigma^{-1} \left( \lat_i - m \right) h_i(\lat_i) \right|} <\infty.
\end{align*}
\end{proof}

\begin{lemma}
\label{lemma:gauss_bound_property_nd_cont}
Let $h(\cdot):\mathcal{R}^d\mapsto \mathcal{R}$ be continuously differentiable and its derivative is locally ACL.
$q(\vlat):=\gauss(\vlat|\vmu,\vSigma)$ is a multivariate Gaussian distribution with mean $\vmu$ and variance $\vSigma$.
Let $\left|\left|\cdot\right|\right|$ denote the Euclidean norm for a vector and the Frobenius norm for a matrix.
If $\Unmyexpect{q}\sqr{\left|\left| \nabla_\lat^2 h(\vlat) \right|\right|}<\infty$,
$\Unmyexpect{q}\sqr{\left|\left| \nabla_\lat h(\vlat) \right|\right|}<\infty$, and 
$\Unmyexpect{q}\sqr{\left| h(\vlat) \right|} <\infty$, then
\begin{align*}
\Unmyexpect{q}\sqr{\left|\left| \vSigma^{-1} (\vlat-\vmu)  \nabla_\lat^T h(\vlat) \right|\right|} <\infty, \quad
\Unmyexpect{q}\sqr{\left|\left| \vSigma^{-1} (\vlat-\vmu)   h(\vlat) \right|\right|} <\infty.
\end{align*}
\end{lemma}
\begin{proof}
We define a scalar function $f_{i}(\vlat):=\nabla_\lat^T h(\vlat) \ve_i$.
Since $\Unmyexpect{q}\sqr{\left|\left| \nabla_\lat f_i(\vlat) \right|\right|}=\Unmyexpect{q}\sqr{\left|\left| \nabla_\lat^2 h(\vlat) \ve_i \right|\right|}<\infty$,
$\Unmyexpect{q}\sqr{\left|f_i(\vlat) \right|}=\Unmyexpect{q}\sqr{\left|  \nabla_\lat^T h(\vlat) \ve_i \right|}<\infty$,
by applying Lemma \ref{lemma:gauss_bound_property_nd} to $f_i(\vlat)$, we have
\begin{align*}
\Unmyexpect{q}\sqr{\left|\left| \vSigma^{-1} (\vlat-\vmu) \left( \nabla_\lat^T (\vlat) \ve_i\right) \right|\right|}=
\Unmyexpect{q}\sqr{\left|\left| \vSigma^{-1} (\vlat-\vmu) f_i(\vlat) \right|\right|} <\infty
\end{align*}
Since index $i$ is arbitrary, we know that
$\Unmyexpect{q}\sqr{\left|\left| \vSigma^{-1} (\vlat-\vmu)  \nabla_\lat^T h(\vlat) \right|\right|} <\infty$.

Likewise, since 
$\Unmyexpect{q}\sqr{\left|\left| \nabla_\lat h(\vlat) \right|\right|}<\infty$, and 
$\Unmyexpect{q}\sqr{\left| h(\vlat) \right|} <\infty$, 
by applying Lemma \ref{lemma:gauss_bound_property_nd} to $h(\vlat)$, we have
\begin{align*}
\Unmyexpect{q}\sqr{\left|\left| \vSigma^{-1} (\vlat-\vmu)   h(\vlat) \right|\right|} <\infty.
\end{align*}
\end{proof}

   \section{Gradient Identities for Univariate Continuous Exponentially-family Distributions}
   \label{app:exp_dist}

\subsection{Proof of Theorem \ref{claim:impl_repm_trick_1d}}

\begin{proof}
It is easy to verify that $\tilde{f_i}(\lat)=f_i(\lat) h(\lat)$ is locally AC since $f_i(\lat)$ and $h(\lat)$ are both locally AC.  By Lemma \ref{lemma:stein_exp_1d}, we have
\begin{align}
 -\Unmyexpect{q} \sqr{ \tilde{f_i}(\lat) \frac{\nabla_{\lat}q(\lat|\vvarpar_\lat)}{q(\lat|\vvarpar_\lat)} } =& \Unmyexpect{q}\sqr{ \nabla_{\lat} \tilde{f_i}(\lat)} \label{eq:imp_rep_trick_1d}
\end{align}

Notice that
\begin{align*}
 \nabla_{\lat} \tilde{f_i}(\lat) =
 -\Big[\underbrace{\frac{  \nabla_{\varpar_{i}} \psi(\lat,\vvarpar_\lat)}{q(\lat|\vvarpar_\lat)}}_{=f_i(\lat)} \nabla_\lat h(\lat) +   \frac{ h(\lat)\nabla_{\varpar_{i}} q(\lat|\vvarpar_\lat)  }{q(\lat|\vvarpar_\lat)} + \frac{ \tilde{f_i}(\lat) \nabla_\lat q(\lat|\vvarpar_\lat) }{q(\lat|\vvarpar_\lat)} \Big] 
\end{align*}
The expression \eqref{eq:imp_rep_trick_1d} can be re-expressed as

\begin{align}
 -\Unmyexpect{q} \Big[  \quad \cancelto{}{\tilde{f_i}(\lat) \frac{\nabla_{\lat}q(\lat|\vvarpar_\lat)}{q(\lat|\vvarpar_\lat)}} \, \quad \Big] =& - \Unmyexpect{q}\Big[ f_i(\lat) \nabla_\lat h(\lat) +   \frac{ h(\lat)\nabla_{\varpar_{i}} q(\lat|\vvarpar_\lat)  }{q(\lat|\vvarpar_\lat)} + \cancelto{}{\frac{ \tilde{f_i}(\lat) \nabla_\lat q(\lat|\vvarpar_\lat) }{q(\lat|\vvarpar_\lat)}}   \quad \Big] \label{eq:imp_rep_trick_1d_eq}
\end{align}

By \eqref{eq:imp_rep_trick_1d_eq}, we have the following identity
\begin{align*}
 \Unmyexpect{q}\sqr{     \frac{ h(\lat)\nabla_{\varpar_{i}} q(\lat|\vvarpar_\lat)  }{q(\lat|\vvarpar_\lat)}    }
= -\Unmyexpect{q}\sqr{f_i(\lat)\nabla_\lat h(\lat) }
\end{align*}

Since we can interchange the integration and differentiation, we know that
\begin{align*}
 \nabla_{\varpar_{i}} \Unmyexpect{q}\sqr{  h(\lat)} = 
\Unmyexpect{q}\sqr{     \frac{ h(\lat)\nabla_{\varpar_{i}} q(\lat|\vvarpar_\lat)  }{q(\lat|\vvarpar_\lat)}    }
= -\Unmyexpect{q}\sqr{f_i(\lat)\nabla_\lat h(\lat) }
\end{align*}

\end{proof}

        \section{Gradient Identities for Gaussian Variance-mean Mixtures}
   \label{app:gvm_dist}
      
\subsection{Proof of Theorem \ref{claim:bonnet_gauss_mixture}}
\begin{proof}
 Let's consider the gradient identity for $\valpha$. By the assumptions, we can the integration and differentiation to obtain the following expression.
 \begin{align*}
 \nabla_{\alpha} \Unmyexpect{q(\lat)} \sqr{ h(\vlat) } 
 &= \int  \nabla_{\alpha} q(\vlat|\vmu,\valpha,\vSigma) h(\vlat) d\vlat \\
 &= \int   \nabla_{\alpha }\sqr{ \int q(\mix,\vlat|\vmu,\valpha,\vSigma)d\mix} h(\vlat) d\vlat \\
 &= \int    \sqr{ \int \frac{ u(\mix)}{ v(\mix)} \vSigma^{-1} \left( \vlat-\vmu-u(\mix)\valpha  \right) q(\mix, \vlat|\vmu,\valpha,\vSigma) d\mix} h(\vlat) d\vlat \\
 &= \Unmyexpect{q(\mix,\lat)}\sqr{ \frac{ u(\mix)}{ v(\mix)} \vSigma^{-1} \left( \vlat-\vmu-u(\mix)\valpha  \right) h(\vlat) }
 \end{align*}
Conditioned on $\mix$, $q(\vlat|\mix)$ is Gaussian denoted by  $q(\vlat|\mix):=\gauss(\vlat|\vmu+u(\mix)\valpha,v(\mix)\vSigma)$.
By applying Lemma \ref{lemma:stein_gauss_mean_nd} to  $u(\mix)h(\vlat)$,
we have
\begin{align*}
  \Unmyexpect{q(\mix,\lat)}\sqr{  \nabla_{\lat} \left(u(\mix) h(\vlat)\right) }   
&=  \Unmyexpect{q(\mix)}\sqr{ \Unmyexpect{q(\lat|\mix)}\sqr{ \nabla_{\lat} \left(u(\mix) h(\vlat)\right) }   } \\
 &= 
 \Unmyexpect{q(\mix)}\sqr{    \Unmyexpect{q(\lat|\mix)} \sqr{ \left(v(\mix) \vSigma \right)^{-1} \left( \vlat-\vmu-u(\mix)\valpha \right) \left(u(\mix) h(\vlat)\right) } }\\
 &=\Unmyexpect{q(\mix,\lat)}\sqr{ \frac{u(\mix)}{v(\mix)}  \vSigma^{-1} \left( \vlat-\vmu-u(\mix)\valpha \right) h(\vlat)  }
\end{align*}

Therefore, we have
\begin{align*}
  \nabla_{\alpha} \Unmyexpect{q(\lat)} \sqr{ h(\vlat) } =
  \Unmyexpect{q(\mix,\lat)}\sqr{ u(\mix) \nabla_{\lat} h(\vlat)}  
\end{align*}

Similarly, we can show that
\begin{align*}
\nabla_{\mu} \Unmyexpect{q(\lat)} \sqr{ h(\vlat) } =  \Unmyexpect{q(\mix,\lat)}\sqr{ \nabla_{\lat} h(\vlat)}  =  \Unmyexpect{q(\lat)}\sqr{ \nabla_{\lat} h(\vlat)}  
\end{align*}

\end{proof}

\subsection{Example \ref{emp:skew_gauss}}
\label{app:emp_skew_gauss}
A concrete example is the multivariate skew Gaussian distribution.
\begin{align*}
q(\mix,\vlat|\vmu,\valpha,\vSigma):= & \gauss(\vlat|\vmu +  |\mix| \valpha, \vSigma) \gauss(\mix|0,1)  \\
 q(\vlat|\vmu,\valpha,\vSigma) := &  \int q(\mix,\vlat|\vmu,\valpha,\vSigma)  d\mix \\
=&  2 \Phi\left( \frac{ \left( \vlat - \vmu \right)^T \vSigma^{-1} \valpha }{ \sqrt{ 1+\valpha^T\vSigma^{-1}\valpha} } \right)\gauss( \vlat| \vmu, \vSigma+\valpha\valpha^T ) 
\end{align*} where $u(\mix)=|\mix|$ and $v(\mix)=1$.

Furthermore, we have
\begin{align*}
&  \int |\mix| q(\mix,\vlat|\vmu,\valpha,\vSigma) d\mix \\
&=  u_1(\vlat,\vmu,\valpha,\vSigma) \gauss( \vlat| \vmu, \vSigma )  + u_2(\vlat,\vmu,\valpha,\vSigma) q(\vlat|\vmu,\valpha,\vSigma) 
\end{align*} where
$u_1(\vlat,\vmu,\valpha,\vSigma)=\frac{\sqrt{2/\pi} }{ 1+\valpha^T \vSigma^{-1} \valpha   }$,
$u_2(\vlat,\vmu,\valpha,\vSigma)=\frac{ \left( \vlat - \vmu \right)^T \vSigma^{-1} \valpha }{1+\valpha^T\vSigma^{-1}\valpha}$.

\subsection{Example \ref{emp:exp_gauss}}
\label{app:emp_exp_gauss}
Another example is the multivariate exponentially modified Gaussian distribution.
\begin{align*}
q(\mix,\vlat|\vmu,\valpha,\vSigma):= & \gauss(\vlat|\vmu +  \mix \valpha, \vSigma) \expdist(\mix|1)  \\
 q(\vlat|\vmu,\valpha,\vSigma) := &  \int_{0}^{+\infty} q(\mix,\vlat|\vmu,\valpha,\vSigma)  d\mix \\
=&  \frac{ \sqrt{2\pi}\mathrm{det}\left(2\pi\vSigma \right)^{-\half} }{  \sqrt{\valpha^T \vSigma^{-1}\valpha} } \Phi\left( \frac{ \left( \vlat-\vmu \right)^T \vSigma^{-1} \valpha -1 }{ \sqrt{ \valpha^T \vSigma^{-1} \valpha }  } \right)\exp\crl{  \half\sqr{ \frac{ \left( \left( \vlat-\vmu \right)^T \vSigma^{-1} \valpha -1 \right)^2  }{ \valpha^T \vSigma^{-1} \valpha } - \left(\vlat-\vmu\right)^T \vSigma^{-1} \left(\vlat-\vmu \right)  }  }
\end{align*} where $u(\mix)=\mix$ and $v(\mix)=1$.

Furthermore, we have
\begin{align}
  \int_{0}^{+\infty} \mix q(\mix,\vlat|\vmu,\valpha,\vSigma) d\mix 
&=  u_1(\vlat,\vmu,\valpha,\vSigma) \gauss( \vlat| \vmu, \vSigma )  + u_2(\vlat,\vmu,\valpha,\vSigma) q(\vlat|\vmu,\valpha,\vSigma) 
\end{align} where
$u_1(\vlat,\vmu,\valpha,\vSigma)=\frac{1}{ \valpha^T\vSigma^{-1}\valpha} $ and
$u_2(\vlat,\vmu,\valpha,\vSigma)=\frac{ \left(\vlat-\vmu\right)^T\vSigma^{-1}\valpha -1 }{ \valpha^T\vSigma^{-1}\valpha }$.

\subsection{Proof of Theorem \ref{claim:price_gauss_mixture}}

\begin{proof}

Firstly, note that
\begin{align*}
 \Unmyexpect{q(\mix,\lat)}\sqr{ v(\mix) \nabla_{\lat}^2  h(\vlat) }
 &= \Unmyexpect{q(\mix)}\sqr{ \Unmyexpect{q(\lat|\mix)} \sqr{ \nabla_{\lat}^2 \left( v(\mix) h(\vlat) \right) } } 
\end{align*}




Conditioned on $\mix$, $q(\vlat|\mix)$ is Gaussian denoted by  $q(\vlat|\mix):=\gauss(\vlat|\vmu+u(\mix)\valpha,v(\mix)\vSigma)$.
By applying Lemma \ref{lemma:stein_gauss_var_price} to $v(\mix)h(\vlat)$, we have
\begin{align}
 \Unmyexpect{q(\mix,\lat)}\sqr{ v(\mix) \nabla_{\lat}^2  h(\vlat) }
 &= \Unmyexpect{q(\mix)}\sqr{ \Unmyexpect{q(\lat|\mix)}\sqr{  \nabla_{\lat}^2 \left(v(\mix) h(\vlat)\right) } } \nonumber \\
&= \Unmyexpect{q(\mix)}\sqr{ \Unmyexpect{q(\lat|\mix)}\sqr{  \left(v(\mix)\vSigma\right)^{-1} \left( \vlat-\vmu-u(\mix)\valpha \right)   \nabla_{\lat}^T \left(v(\mix) h(\vlat)\right) } } \nonumber \\
&= \Unmyexpect{q(\mix)}\sqr{ \Unmyexpect{q(\lat|\mix)}\sqr{  \vSigma^{-1} \left( \vlat-\vmu-u(\mix)\valpha \right)   \nabla_{\lat}^T  h(\vlat) } }\label{eq:smg_eq3}
\end{align}

By applying Lemma \ref{lemma:stein_gauss_var_reparam} to $v(\mix) h(\vlat)$,
we have the following expression.
\begin{align}
 &\Unmyexpect{q(\mix,\lat) }\sqr{ \vSigma^{-1}  \sqr{v^{-1}(\mix) \left( \vlat-\vmu - u(\mix)\valpha \right)  \left( \vlat-\vmu - u(\mix)\valpha \right)^T - \vSigma } \vSigma^{-1}   h(\vlat) } \nonumber \\
= &\Unmyexpect{q(\mix)} \sqr{\Unmyexpect{q(\lat|\mix)} \sqr{\left( v(\mix) \vSigma\right)^{-1}  \sqr{\left( \vlat-\vmu - u(\mix)\valpha \right)  \left( \vlat-\vmu - u(\mix)\valpha \right)^T - v(\mix)\vSigma } \left(v(\mix)\vSigma\right)^{-1}  \left(v(\mix) h(\vlat)\right) } } \nonumber\\
= &\Unmyexpect{q(\mix)} \sqr{\Unmyexpect{q(\lat|\mix)} \sqr{\left( v(\mix)\vSigma \right)^{-1}  \left( \vlat-\vmu-u(\mix)\valpha  \right)   \nabla_{\lat}^T \left( v(\mix)  h(\vlat) \right) } } \nonumber\\
=& \Unmyexpect{q(\mix,\lat)}\sqr{  \vSigma^{-1} \left( \vlat-\vmu - u(\mix)\valpha \right)    \nabla_{\lat}^T h(\vlat) }  \label{eq:smg_eq4}
\end{align}



By the regular assumptions, we can swap the integration and differentiation to get the following expression.
\begin{align}
\nabla_{\Sigma} \Unmyexpect{q(\lat)} \sqr{ h(\vlat) } 
&= \int \nabla_{\Sigma} q(\vlat|\vmu,\vSigma) h(\vlat) d\vlat \nonumber\\
&= \int \int \sqr{ \nabla_{\Sigma} \gauss(\vlat|\vmu,v(\mix)\vSigma) q(\mix) d\mix } h(\vlat) d\vlat \nonumber\\
&= \half \Unmyexpect{q(\mix,\lat)}\sqr{  \vSigma^{-1}  \sqr{ v^{-1}(\mix) \left( \vlat-\vmu -u(\mix)\valpha\right)  \left( \vlat-\vmu -u(\mix)\valpha\right)^T - \vSigma } \vSigma^{-1}  h(\vlat)  } \label{eq:smg_eq5} 
\end{align}



Finally, by Eq. \eqref{eq:smg_eq3} , \eqref{eq:smg_eq4}, and \eqref{eq:smg_eq5}, we have
\begin{align*}
\nabla_{\Sigma} \Unmyexpect{q(\lat)} \sqr{ h(\vlat) }  
= \half \Unmyexpect{q(\mix,\lat)}\sqr{   \vSigma^{-1} \left( \vlat-\vmu -u(\mix)\valpha\right)    \nabla_{\lat}^T h(\vlat) }
= \half \Unmyexpect{q(\mix,\lat)}\sqr{ v(\mix) \nabla_{\lat}^2  h(\vlat) } 
\end{align*}

\end{proof}

\subsection{Example \ref{emp:stu_t}}
\label{app:emp_stu_t}

A concrete example is the multivariate Student's t-distribution with fixed degree of freedom $2\beta$.
We consider a case when $\beta>1$, since the variance does not exist when $\beta\leq 1$.
\begin{align*}
q(\mix,\vlat|\vmu,\valpha,\vSigma):= & \gauss(\vlat|\vmu ,\mix \vSigma) \IG(\mix|\beta,\beta)  \\
 q(\vlat|\vmu,\valpha,\vSigma) := &  \int q(\mix,\vlat|\vmu,\valpha,\vSigma)  d\mix \\
=&  \mathrm{det}\left(\pi \vSigma \right)^{-1/2}  \frac{   \Gamma(\beta+d/2)  \left(   2\beta + \left(\vlat-\vmu\right)^T \vSigma^{-1}  \left(\vlat-\vmu\right)    \right)^{-\beta-d/2} }{   \Gamma(\beta)     \left( 2\beta\right)^{-\beta} } .
\end{align*} where $u(\mix)=0$ and $v(\mix)=\mix>0$ since $\mix$ is generate from inverse Gamma distribution $\IG(\mix|\beta,\beta)$.

When $\beta>1$,  we have
\begin{align}
  \int \mix q(\mix,\vlat|\vmu,\valpha,\vSigma) d\mix 
&=  v_1(\vlat,\vmu,\valpha,\vSigma) q(\vlat|\vmu,\valpha,\vSigma) 
\end{align} where
$v_1(\vlat,\vmu,\valpha,\vSigma):=\frac{\beta}{(\beta+d/2-1)} \left(1 + (2\beta)^{-1} \left( \vlat-\vmu \right)^T \vSigma^{-1} \left( \vlat-\vmu \right) \right)$.

\subsection{Example \ref{emp:ninv_gauss}}
\label{app:emp_ninv_gauss}
Another example is the multivariate normal inverse-Gaussian distribution, where $\beta>0$ is fixed.
\begin{align*}
q(\mix,\vlat|\vmu,\valpha,\vSigma):= & \gauss(\vlat|\vmu + \mix\valpha ,\mix \vSigma) \IGauss(\mix|1,\beta)  \\
 q(\vlat|\vmu,\valpha,\vSigma) := &  \int q(\mix,\vlat|\vmu,\valpha,\vSigma)  d\mix \\
 =&\frac{\beta^{\half}}{(2\pi)^{\frac{d+1}{2}}} \mathrm{det}\left(  \vSigma \right)^{-1/2} \exp\sqr{ \left(\vlat-\vmu \right)^T \vSigma^{-1} \valpha +\beta } \frac{ 2 \mathcal{K}_{\frac{d+1}{2}} \left( \sqrt{ \left( \valpha^T \vSigma^{-1} \valpha  + \beta \right) \left( \left(\vlat-\vmu \right)^T \vSigma^{-1} \left(\vlat-\vmu \right) + \beta \right) } \right) }{ \left(\sqrt{  \frac{ \valpha^T \vSigma^{-1} \valpha + \beta }{ \left( \vlat-\vmu \right)^T \vSigma^{-1} \left( \vlat-\vmu \right) + \beta } } \right)^{\frac{-d-1}{2}} } 
\end{align*} where $u(\mix)=v(\mix)=\mix>0$ since $\mix$ is generate from an inverse Gaussian distribution 
$\IGauss(\mix|1,\beta)=\left(\frac{\beta}{2\pi \mix^3} \right)^{\half}\exp\crl{ -\frac{\beta}{2}\left(\mix+\mix^{-1}\right) + \beta }$

We have
\begin{align}
  \int \mix q(\mix,\vlat|\vmu,\valpha,\vSigma) d\mix 
&=  v_1(\vlat,\vmu,\valpha,\vSigma) q(\vlat|\vmu,\valpha,\vSigma) 
\end{align} where
\begin{align*}
v_1(\vlat,\vmu,\valpha,\vSigma):= 
\sqrt{  \frac{ \left( \vlat-\vmu \right)^T \vSigma^{-1} \left( \vlat-\vmu \right) + \beta }{ \valpha^T \vSigma^{-1} \valpha + \beta } }  \frac{ \mathcal{K}_{\frac{d-1}{2}} \left( \sqrt{ \left( \valpha^T \vSigma^{-1} \valpha  + \beta \right) \left( \left(\vlat-\vmu \right)^T \vSigma^{-1} \left(\vlat-\vmu \right) + \beta \right) } \right)} { \mathcal{K}_{\frac{d+1}{2}} \left( \sqrt{ \left( \valpha^T \vSigma^{-1} \valpha  + \beta \right) \left( \left(\vlat-\vmu \right)^T \vSigma^{-1} \left(\vlat-\vmu \right) + \beta \right) } \right)} 
\end{align*}
  
         \section{Gradient Identities for Continuous Exponential-family Mixtures}
   \label{app:expm_dist}
      \subsection{Proof of Theorem \ref{claim:impl_repm_trick_2d}}
\label{app:impl_repm_trick_2d_proof}
\begin{proof}

It is easy to verify that $\tilde{f}_{i,j}(\lat_j):= f_{i,j}(\lat_j,\lat_{-j})  h(\lat_j,\lat_{-j}) \prod_{k\geq j+1} q(\lat_k| \vlat_{1:(k-1)},\vvarpar )$ is locally AC since $h(\lat_j,\lat_{-j})$,  $f_{i,j}(\lat_j,\lat_{-j})$, and $q(\lat_k| \vlat_{1:(k-1)},\vvarpar )$ are all locally AC w.r.t. $\lat_j$ for almost every $\lat_{-j}$.

Recall that $f_{i,j}(\vlat) := \ve_j^T \sqr{ \nabla_{\lat} \vPsi(\vlat,\vvarpar) }^{-1} \nabla_{\varpar_i}\vPsi(\vlat,\vvarpar)$.
By the definitions, we have
\begin{align*}
 \left( \nabla_{\lat}\vPsi(\vlat,\vvarpar) \right)^{-1} &= \begin{bmatrix}
\frac{1}{q(\lat_1|\boldsymbol{\varpar})} \, \,& 0 \\
\quad \\
-\frac{ \nabla_{\lat_1}\psi_2(\lat_1,\lat_2,\boldsymbol{\varpar})} {q(\lat_1|\boldsymbol{\varpar})q(\lat_2|\lat_1,\boldsymbol{\varpar})} \,  \, & \frac{1}{q(\lat_2|\lat_1,\boldsymbol{\varpar})}
\end{bmatrix} \\
\\
f_{i,1}(\vlat) &= \frac{ \nabla_{\varpar_i} \psi_1(\lat_1,\boldsymbol{\varpar}) }{ q(\lat_1|\boldsymbol{\varpar})} \\
f_{i,2}(\vlat) &=  -\frac{ f_{i,1}(\vlat)\nabla_{\lat_1}\psi_2(\lat_1,\lat_2,\vvarpar)} {q(\lat_2|\lat_1,\boldsymbol{\varpar})} +  \frac{ \nabla_{\varpar_i} \psi_2(\lat_1,\lat_2,\boldsymbol{\varpar}) }{ q(\lat_2|\lat_1,\boldsymbol{\varpar})} \\
\end{align*}

Note that the following expression holds almost everywhere due to the product rule for locally AC functions.
\begin{align}
 \nabla_{\lat_1} f_{i,1}(\vlat) &=  \frac{\nabla_{\varpar_i} q(\lat_1|\vvarpar) }{q(\lat_1|\vvarpar)}  - f_{i,1}(\vlat)\frac{\nabla_{\lat_1} q(\lat_1|\vvarpar) }{q(\lat_1|\vvarpar)}     \label{eq:impl_repm_trick_l1_eq} 
\end{align}

Similarly, the following expression holds for almost every $\lat_1$ due to the product rule for locally AC functions.
 \begin{align}
 \nabla_{\lat_2} f_{i,2}(\vlat) &= 
 - \frac{ f_{i,1}(\vlat)\nabla_{\lat_1}q(\lat_2|\lat_1,\vvarpar)} {q(\lat_2|\lat_1,\boldsymbol{\varpar})} - f_{i,2}(\vlat) \frac{ \nabla_{\lat_2}(\lat_2|\lat_1,\vvarpar)  }{ q(\lat_2|\lat_1,\vvarpar) } +  \frac{\nabla_{\varpar_i} q(\lat_2|\lat_1,\vvarpar)}{q(\lat_2|\lat_1,\vvarpar) }
   \label{eq:impl_repm_trick_l2_eq} 
\end{align}

Recall that
$\tilde{f}_{i,1}(\lat_1)= f_{i,1}(\vlat)  h(\vlat) q(\lat_2|\lat_1,\vvarpar)$.

By Lemma \ref{lemma:stein_exp_1d}, we have
 \begin{align*}
 & - \Unmyexpect{q(\lat_1)}\big[ \quad \cancelto{}{ q(\lat_2|\lat_1,\vvarpar) h(\vlat)f_{i,1}(\vlat) \frac{ \nabla_{\lat_1} q(\lat_1|\vvarpar) }{ q(\lat_1|\vvarpar) }} \quad \big] \\
  =& -\Unmyexpect{q(\lat_1)}\sqr{\tilde{f}_{i,1}(\lat_1)\frac{\nabla_{\lat_1}q(\lat_1|\vvarpar)}{q(\lat_1|\vvarpar)} }\\
  =& \Unmyexpect{q(\lat_1)}\sqr{ \nabla_{z_1}\tilde{f}_{i,1}(\lat_1)} \\
  =&\Unmyexpect{q(\lat_1)}\sqr{ \nabla_{\lat_1} \sqr{  q(\lat_2|\lat_1,\vvarpar) h(\vlat)f_{i,1}(\vlat) }  } \\
  =& \Unmyexpect{q(\lat_1)}\sqr{ h(\vlat)f_{i,1}(\vlat) \nabla_{\lat_1} q(\lat_2|\lat_1,\vvarpar) + q(\lat_2|\lat_1,\vvarpar) f_{i,1}(\vlat) \nabla_{\lat_1} h(\vlat) + q(\lat_2|\lat_1,\vvarpar) h(\vlat) \nabla_{\lat_1} f_{i,1}(\vlat) } \\
  =& \Unmyexpect{q(\lat_1)}\big[ h(\vlat)f_{i,1}(\vlat) \nabla_{\lat_1} q(\lat_2|\lat_1,\vvarpar) + q(\lat_2|\lat_1,\vvarpar) f_{i,1}(\vlat) \nabla_{\lat_1} h(\vlat) + q(\lat_2|\lat_1,\vvarpar) h(\vlat) \big(\frac{\nabla_{\varpar_i} q(\lat_1|\vvarpar) }{q(\lat_1|\vvarpar)} \cancelto{}{ - f_{i,1}(\vlat)\frac{\nabla_{\lat_1} q(\lat_1|\vvarpar) }{q(\lat_1|\vvarpar)}    } \big)  \big]
 \end{align*} where we obtain the last equation by \eqref{eq:impl_repm_trick_l1_eq}.

 The above expression gives the following identity.
 \begin{align}
 0 &=  
   \Unmyexpect{q(\lat_1)}\sqr{ h(\vlat)f_{i,1}(\vlat) \nabla_{\lat_1} q(\lat_2|\lat_1,\vvarpar) + q(\lat_2|\lat_1,\vvarpar) f_{i,1}(\vlat) \nabla_{\lat_1} h(\vlat) + q(\lat_2|\lat_1,\vvarpar) h(\vlat) \frac{\nabla_{\varpar_i} q(\lat_1|\vvarpar) }{q(\lat_1|\vvarpar)}    }\nonumber \\
   &= \Unmyexpect{q(\lat_1)}\sqr{ h(\vlat)f_{i,1}(\vlat) \nabla_{\lat_1} q(\lat_2|\lat_1,\vvarpar) }+ \Unmyexpect{q(\lat_1,\lat_2)}\sqr{ f_{i,1}(\vlat) \nabla_{\lat_1} h(\vlat) + h(\vlat) \frac{\nabla_{\varpar_i} q(\lat_1|\vvarpar) }{q(\lat_1|\vvarpar)}} \nonumber \\
    &= \Unmyexpect{q(\lat_1,\lat_2)}\sqr{ h(\vlat)f_{i,1}(\vlat)\frac{ \nabla_{\lat_1} q(\lat_2|\lat_1,\vvarpar)}{ q(\lat_2|\lat_1,\vvarpar)} }+ \Unmyexpect{q(\lat_1,\lat_2)}\sqr{ f_{i,1}(\vlat) \nabla_{\lat_1} h(\vlat) + h(\vlat) \frac{\nabla_{\varpar_i} q(\lat_1|\vvarpar) }{q(\lat_1|\vvarpar)}}  \label{eq:impl_repm_trick_l1_final_eq}
 \end{align}

Note that
$\tilde{f}_{i,2}(\lat_1)= f_{i,2}(\vlat)  h(\vlat)$.
Similarly, by Lemma \ref{lemma:stein_exp_1d}, we have
\begin{align*}
 & - \Unmyexpect{q(\lat_1)}\big[\Unmyexpect{q(\lat_2|\lat_1)}\big[  \cancelto{}{  h(\vlat)f_{i,2}(\vlat) \frac{ \nabla_{\lat_2}(\lat_2|\lat_1,\vvarpar)  }{ q(\lat_2|\lat_1,\vvarpar) } }   \big] \big] \\
 =& -\Unmyexpect{q(\lat_1)} \Unmyexpect{q(\lat_2|\lat_1)} \sqr{ \tilde{f}_{i,2}(\lat_1)\frac{\nabla_{\lat_2}q(\lat_2|\lat_1)}{q(\lat_2|\lat_1)} }\\
 =& \Unmyexpect{q(\lat_1)} \Unmyexpect{q(\lat_2|\lat_1)} \sqr{ \nabla_{\lat_2}\tilde{f}_{i,2}(\lat_1) }\\
 =&  \Unmyexpect{q(\lat_1)}\sqr{\Unmyexpect{q(\lat_2|\lat_1)}\sqr{  \nabla_{\lat_2}\sqr{ h(\vlat)f_{i,2}(\vlat)}  } } \\
 =& \Unmyexpect{q(\lat_1)}\sqr{\Unmyexpect{q(\lat_2|\lat_1)}\sqr{ f_{i,2}(\vlat) \nabla_{\lat_2}h(\vlat) + h(\vlat) \nabla_{\lat_2} f_{i,2}(\vlat)} } \\
 =& \Unmyexpect{q(\lat_1)}\big[ \Unmyexpect{q(\lat_2|\lat_1)}\big[ f_{i,2}(\vlat) \nabla_{\lat_2}h(\vlat) + h(\vlat) \big(      
 - \frac{ f_{i,1}(\vlat)\nabla_{\lat_1}q(\lat_2|\lat_1,\vvarpar)} {q(\lat_2|\lat_1,\boldsymbol{\varpar})} \cancelto{}{- f_{i,2}(\vlat) \frac{ \nabla_{\lat_2}(\lat_2|\lat_1,\vvarpar)  }{ q(\lat_2|\lat_1,\vvarpar) } } +  \frac{\nabla_{\varpar_i} q(\lat_2|\lat_1,\vvarpar)}{q(\lat_2|\lat_1,\vvarpar) } \big) \big]  \big]
 \end{align*} where we obtain the last equation by \eqref{eq:impl_repm_trick_l2_eq}.

The above expression gives the following identity. 
\begin{align}
 0 &= \Unmyexpect{q(\lat_1)q(\lat_2|\lat_1)}\sqr{ f_{i,2}(\vlat) \nabla_{\lat_2}h(\vlat) + h(\vlat) \left(      
 \frac{ -f_{i,1}(\vlat)\nabla_{\lat_1}q(\lat_2|\lat_1,\vvarpar)} {q(\lat_2|\lat_1,\boldsymbol{\varpar})} +  \frac{\nabla_{\varpar_i} q(\lat_2|\lat_1,\vvarpar)}{q(\lat_2|\lat_1,\vvarpar) } \right) } \label{eq:impl_repm_trick_l2_final_eq}
\end{align}

By
\eqref{eq:impl_repm_trick_l1_final_eq} and
\eqref{eq:impl_repm_trick_l2_final_eq}, we have
\begin{align}
 0 &= \Unmyexpect{q(\lat_1)q(\lat_2|\lat_1)}\sqr{ f_{i,1}(\vlat) \nabla_{\lat_1}h(\vlat) + f_{i,2}(\vlat) \nabla_{\lat_2}h(\vlat) + h(\vlat) \left(  \frac{\nabla_{\varpar_i} q(\lat_2|\lat_1,\vvarpar)}{q(\lat_2|\lat_1,\vvarpar) }  + \frac{ \nabla_{\varpar_i} q(\lat_1|\vvarpar)}{ q(\lat_1|\vvarpar) }  \right) } \label{eq:imp_rep_trick_2d_eq}
\end{align}

Therefore, by \eqref{eq:imp_rep_trick_2d_eq}, we have  the following identity
\begin{align*}
  \Unmyexpect{q(\lat_1,\lat_2)}\sqr{  h(\vlat) \left(  \frac{\nabla_{\varpar_i} q(\lat_2|\lat_1,\vvarpar)}{q(\lat_2|\lat_1,\vvarpar) }  + \frac{ \nabla_{\varpar_i} q(\lat_1|\vvarpar)}{ q(\lat_1|\vvarpar) }  \right) }
 =  -\Unmyexpect{q(\lat_1,\lat_2)}\sqr{  f_{i,1}(\vlat) \nabla_{\lat_1}h(\vlat) + f_{i,2}(\vlat) \nabla_{\lat_2}h(\vlat)}
\end{align*}
Since we can interchange the integration and differentiation, we know that
\begin{align*}
 \nabla_{\varpar_{i}} \Unmyexpect{q}\sqr{  h(\vlat)} = 
 \Unmyexpect{q}\sqr{    h(\vlat) \left(  \frac{ \nabla_{\varpar_{i}} q(\lat_1|\vvarpar_\lat)  }{q(\lat_1|\vvarpar_\lat)}  +  \frac{ \nabla_{\varpar_{i}} q(\lat_2|\lat_1,\vvarpar_\lat)  }{q(\lat_2|\lat_1,\vvarpar_\lat)}\right)    }
 = -\Unmyexpect{q(\lat_1,\lat_2)}\sqr{  f_{i,1}(\vlat) \nabla_{\lat_1}h(\vlat) + f_{i,2}(\vlat) \nabla_{\lat_2}h(\vlat)}
 ,
\end{align*} which gives the desired identity.
\end{proof}

\end{appendix}

\end{document}